\documentclass[10pt]{article} 
\usepackage[preprint]{tmlr}

\usepackage{amsmath,amsfonts,bm}

\def\eqref#1{equation~\ref{#1}}

\def\1{\bm{1}}

\DeclareMathAlphabet{\mathsfit}{\encodingdefault}{\sfdefault}{m}{sl}
\SetMathAlphabet{\mathsfit}{bold}{\encodingdefault}{\sfdefault}{bx}{n}

\newcommand{\R}{\mathbb{R}}

\DeclareMathOperator*{\argmin}{arg\,min}

\usepackage{url}

\usepackage{microtype}
\usepackage{graphicx}
\usepackage{subcaption}
\usepackage{booktabs} 

\usepackage{sidecap}
\sidecaptionvpos{figure}{t}

\usepackage{amsmath, amsfonts, amssymb}
\usepackage{mathtools}
\usepackage{amsthm}
\usepackage{nicefrac}       
\usepackage{microtype}      
\usepackage[ruled,vlined,algo2e]{algorithm2e}

\usepackage[table]{xcolor}
\usepackage{booktabs}
\usepackage{tabularx}
\usepackage{array}

\definecolor{PastelBlueBg}{HTML}{EBF3F9}     
\definecolor{PastelBlueFrame}{HTML}{A9C7DF}  

\usepackage{pifont}   
\usepackage{multirow} 

\usepackage{float}

\usepackage{dblfloatfix}
\usepackage{tikz}

\usepackage{enumitem}

\usepackage[most]{tcolorbox}  

\usepackage[normalem]{ulem}   

\usepackage{hyperref}
\usepackage[capitalize,noabbrev]{cleveref}

\definecolor{purplepastel}{RGB}{160, 130, 180}
\definecolor{bluepastel}{RGB}{120, 150, 190}

\newcommand{\D}{\mathcal{D}}

\newcommand{\T}{\mathcal{T}}
\newcommand{\Sc}{\mathcal{S}}

\newcommand{\x}{\>x}
\newcommand{\y}{\>y}

\newcommand\egaldef{\stackrel{\mbox{\upshape\tiny def}}{=}}

\newcommand{\ind}[1]{\mathbb{1}_{\{#1\}}}
\newcommand\EE{\mathsf{E}}

\def\DD{\displaystyle}

\definecolor{darkgreen}{RGB}{26, 148, 49}
\newcommand{\cf}[1]{\textcolor{black}{#1}}

\newcolumntype{L}[1]{>{\raggedright\arraybackslash}p{#1}}

\theoremstyle{plain}

\theoremstyle{definition}

\theoremstyle{remark}

\usepackage[textsize=tiny]{todonotes}

\title{PRIVET: PRoximIty leakage detection Via Extreme value Theory}

\author{
\normalfont
\parbox{\textwidth}{\centering
\textbf{Antoine Szatkownik}$^{1}$, 
\textbf{Aurélien Decelle}$^{1,2,3}$, 
\textbf{Beatriz Seoane}$^{2}$, 
\textbf{Nicolas Béreux}$^{1}$, 
\textbf{Léo Planche}$^{1}$, 
\textbf{Guillaume Charpiat}$^{1}$, 
\textbf{Burak Yelmen}$^{4}$, 
\textbf{Flora Jay}$^{1}$, 
\textbf{Cyril Furtlehner}$^{1}$\\[6pt]
$^{1}$Université Paris-Saclay, CNRS, INRIA, LISN, Gif-sur-Yvette, France\\
$^{2}$Departamento de Física Teórica, Universidad Complutense de Madrid, Madrid, Spain\\
$^{3}$Escuela Técnica Superior de Ingenieros Industriales, Universidad Politécnica de Madrid, Madrid, Spain\\
$^{4}$Estonian Genome Centre, Institute of Genomics, University of Tartu, Tartu, Estonia\\[3pt]
\texttt{antoine.szatkownik@gmail.com}
}
}

\def\month{MM}  
\def\year{YYYY} 
\def\openreview{\url{https://openreview.net/forum?id=XXXX}} 

\begin{document}

\maketitle

\begin{abstract}
Deep generative models are often trained on sensitive data, such as genetic sequences, health data, or more broadly, any copyrighted, licensed or protected content. This raises critical concerns around privacy-preserving synthetic data, and more specifically around privacy leakage, an issue closely tied to overfitting. \cf{Existing proximity-based methods mostly assess privacy risk through global criteria, which quantify a model's overall behaviour but cannot attribute risk to an individual record. Sample-level outputs do exist but they are either uncalibrated, discontinuous, or blind to leakage occurring while the model is globally underfit, which limits their practical use.}
Using extreme value statistics on nearest-neighbor distances, we propose PRIVET, a generic sample-based, modality-agnostic algorithm that assigns an individual  \cf{proximity} leak score to each synthetic sample. \cf{These are evaluated under a chosen representation and distance, each synthetic sample being assigned a continuous score measuring how improbable its proximity to the training set is under a no-leakage model. We empirically demonstrate that PRIVET detects memorization and more subtle forms of proximity-based data leakage across diverse data modalities, including settings with very high dimensionality and limited sample sizes such as genetic data, and in underfitting regimes that overfitting-based diagnostics cannot reach by construction. 
Our analysis further shows that the representation bounds what any distance-based evaluation can detect, existing computer vision embeddings failing to yield perceptually meaningful distances for near-duplicate samples. Accordingly, a low score is evidence of leakage in the chosen metric, while its absence is not a certificate of privacy.}
\end{abstract}

\section{Introduction}
\subsection{Position of the problem}

In the context of unsupervised learning, recent advances in deep learning have enabled the development of generative models capable of producing highly realistic artificial data from a collection of examples. Prominent examples include Generative Adversarial Networks (GANs)\citet{goodfellow2014generative}, Variational Autoencoders (VAEs) \citet{kingma2014auto}, normalizing flows~\citet{rezende2015variational}, diffusion models~\citet{ho2020denoising}, flow matching techniques~\citet{liu2022flow}, energy-based models~\citet{lecun2006tutorial} or Large Language Models (LLMs) ~\citep{attentionisallyouneed}. When applied to image data, these models can synthesize remarkably lifelike samples -- for instance, photorealistic human faces that do not correspond to real individuals. One major difficulty is to assess the quality of these models. A widely used metric is the Fréchet Inception Distance (FID) score~\citet{heusel2017gans}, 
allows one to assess the visual quality and diversity of the generated samples, but is not specifically designed to detect overfitting or privacy leakage and often fails to do so \citep{NEURIPS2023_68b13860}. 
Despite their high capacity to imitate real data distributions, generative models are prone to overfitting due to their typically large number of trainable parameters, particularly when trained on limited datasets. This vulnerability affects many applications and becomes particularly critical in domains such as genomics, where the training data may contain identifiable and sensible information -- for example, genetic sequences associated with specific diseases.
As discussed in~\citet{Yeom_2018}, privacy leakage and overfitting are closely related phenomena, yet conceptually distinct. Overfitting may facilitate privacy attacks, but privacy leaks can occur without it. Multiple forms of privacy leakage have been identified in the literature, the most commonly studied being {\em membership inference} and {\em attribute inference}. Membership inference aims to determine whether a specific sample was used during training, while attribute inference seeks to reconstruct sensitive attributes of individuals from model outputs. Additional privacy threats include model inversion and influence-based attacks, applicable under varying levels of adversarial access. These attacks relate to broader notions of stability in machine learning (ML), and raise serious concerns in domains involving sensitive data, such as genomics. 
Privacy leaks in genetic data were identified early on~\citet{HSRDTMPSNC_2008,SaObJoHa}, with recent work introducing formal metrics and empirical evaluations \citep{OGC_2021, Paige2021-bm}. \citet{berger_privacy} developed a way to detect leakage at the sample level by only looking at summary statistics of genetic variation data. However, summary statistics rely on handcrafted transformations of the data, often involving expert knowledge; hence, such approaches are domain-specific.  \textcolor{black}{Instead, we aim to develop a model-free,  domain-agnostic framework, complementary to 
model-access attacks, centered on the following question: 
\emph{given query and reference datasets, when is a 
query sample anomalously close to the reference?} Our primary focus is on proximity-based privacy leakage on the outputs of a generative model rather than on the upstream generating process, though the framework applies more broadly to any query dataset. For a wider overview of privacy  challenges in ML, see~\citep{DELGROSSO20231}.}
\subsection{Related work} \label{sec:related_work}

Evaluating synthetic data is a critical step that extends beyond the development of generative models. Such data must satisfy key desiderata, including high quality -- capturing both fidelity and diversity -- and {\em novelty}, i.e. no  privacy leakage or memorization (i.e. regurgitation of training samples).

Among existing evaluation metrics, the holistic approaches aim to broadly capture all these aspects simultaneously. For example, \citet{NEURIPS2023_68b13860} introduced a score that correlates with fidelity and diversity, and can also be adapted to assess memorization and overfitting. In contrast, property-specific metrics focus on evaluating a particular aspect of the data. Our proposed method aligns with that, offering a granular assessment of proximity-based privacy leakage at the sample level, which can further be extended to a global measure.

\textbf{Metrics at the dataset-level.} In the context of sensitive data,~\citet{YALE_2020} introduced the Nearest Neighbor Adversarial Accuracy ($\mathcal{AATS}$), a score in $[0,1]$ estimating the resemblance between generated and real training data. A value below $0.5$ indicates overfitting, while above $0.5$ suggests underfitting. This score can also be adapted into a privacy metric by comparing its value to that obtained from a separate test set. $\mathcal{AATS}$ requires equal sample sizes between training and synthetic data, which can be limiting, especially when test sets are significantly smaller. In that vein, $C_T$~\citet{meehanNonParametricTestDetect2020} is a statistical test to detect anomalous resemblance of generated data to training ones. More recently, ~\citet{NEURIPS2023_68b13860, lemos2025pqmass} proposed holistic approaches, Feature Likelihood Divergence (FLD) and PQMass, that rely on kernel density estimation and statistical tests on counts of data points falling in cells constructed by a tesselation of space. These approaches have important limitations: they fail in low-data regimes or produce very noisy estimates, and lack interpretability, \textcolor{black}{i.e., they output continuous scores that neither identify which samples are leaked nor convey how many individuals are affected.} Such shortcomings hinder their use in downstream tasks like sample selection, despite the need for precise, transparent metrics in privacy evaluation—an issue underscored by regulations such as the GDPR~\citep{gdpr2016}.

\textbf{Metrics at the sample-level.}
Like $\mathcal{AATS}$, the Authenticity score~\citet{pmlr-v162-alaa22a} uses nearest-neighbor (NN) distances and provides an interpretable ``leak'' or ``no-leak'' label for each synthetic sample, but leaves aside statistical aspects, making it vulnerable to fluctuations where a synthetic sample may seem close to a training point by chance. Indeed, different densities between train and synthetic set due to different dataset sizes will cause the failure of this method. Moreover, by construction, In-Auth cannot detect privacy leakage in the underfitting regime.
FLD can additionally be used to assign a real number (cf.~definition 3.1 in~\citet{NEURIPS2023_68b13860}) at the sample level, but this score hasn't been thoroughly tested and still requires manual inspection to confirm privacy breaches.

\subsection{Main contributions}

\begin{figure*}[!h]
\centering
\includegraphics[width=1\textwidth]{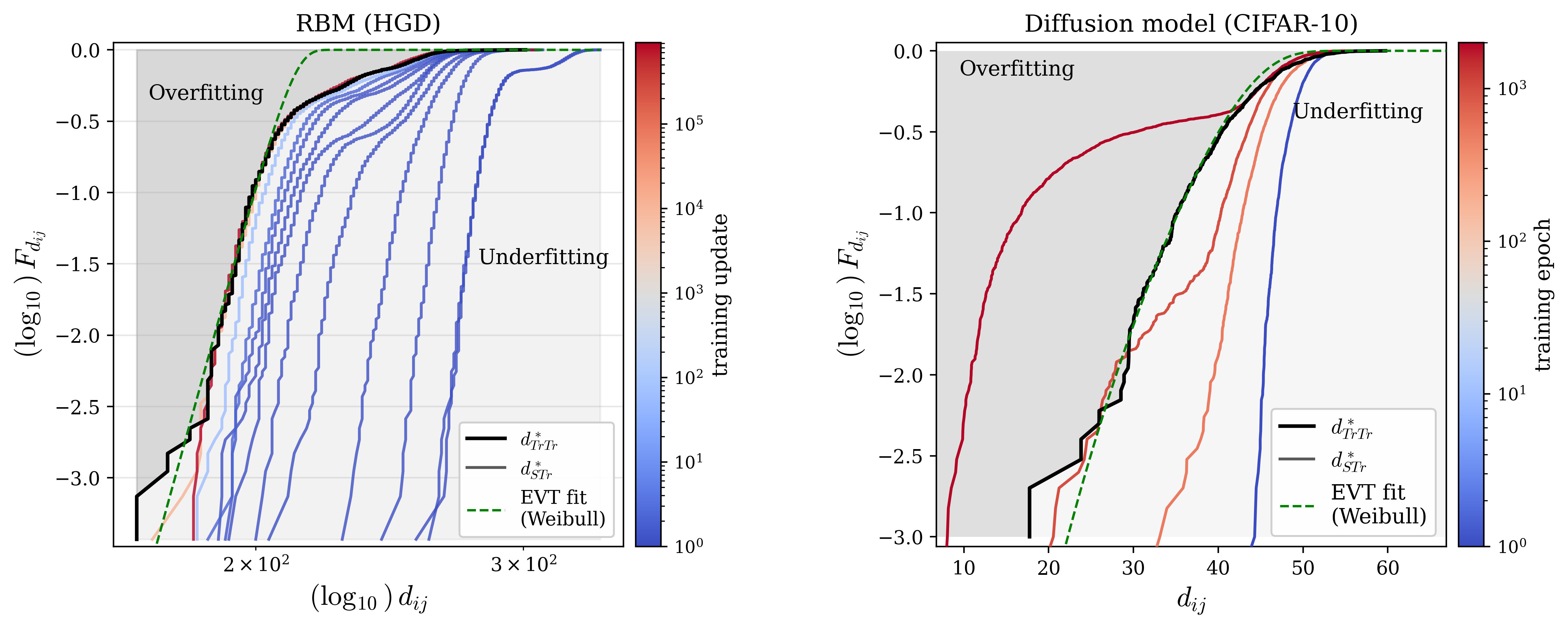}
\caption{\cf{\textbf{Evolution of the  distribution of nearest-neighbor (NN) distances between synthetic data and training data as a function of the number of epochs.} The black curve is the reference train-train NN distance distribution fitted in dotted green by an extreme value distribution {\bf separating the overfitting domain from the underfitting one}. The left panel corresponds to an ideal learning process (performed with a restricted Boltzmann machine on the human genome dataset~\citet{bereux2025fast}) where at some point both reference and synthetic to train distribution coincide. On the right panel instead we observe that no perfect learning (performed with a ad-hoc diffusion model on CIFAR-10) is reached, because some region of the training set start to be overfitted while other region are still underfitted}} 
\label{fig:leak_CDF}
\end{figure*}

Our contributions can be summarized as follows:
\cf{
\begin{itemize}[noitemsep, topsep=0pt]
\item We observe that NN tail distribution constitute a distinct marker of over/under fitting of generative models (see Figure~\ref{fig:leak_CDF}), and empirically demonstrate, across a wide range of datasets and modalities, that extreme value fits 
applied on NN tails distribution are reliable and stable. 
\item Using these fits we build a proximity index based on $r$-th order NN statistics which can be turn into a global or an individual assessment  of generated samples quality regarding over/under-fitting.  
\item We empirically show that this index can be used to detect some form of proximity-based data leakage when the generative model underfit the data. 
\end{itemize}
}
\cf{Note that this assessment rely on a representation and a distance chosen by the practitioner.
As a consequence the evaluation is conditional on that choice, which means that a bad score is evidence of  proximity-leakage while a good one does not warrant absence of leakage.}


We tested our method extensively and compared it to existing ones ($\mathcal{AA_{TS}}$ \citet{YALE2020244}, $C_T$ \citet{meehanNonParametricTestDetect2020}, $Auth$ \citet{pmlr-v162-alaa22a}, $FLD$ \citet{NEURIPS2023_68b13860}, $PQM$ \citet{lemos2025pqmass}) using different datasets from different domains, namely genetics and computer vision, where privacy preservation can be crucial. Furthermore, we considered two types of generative models -- RBMs, and diffusion models -- to demonstrate the applicability and robustness of our privacy assessment framework across diverse generative paradigms.

\cf{The paper is organized as follows: in Section~\ref{sec:method} we present empirical observations of NN distributions, which we analyze in terms of extreme value statistics in order to define some statistically calibrated proximity-based leakage scores. Section~\ref{sec:GoF_analysis} we present an empirical study of the robustness of the fit based on extreme value statistics on which the method relies. In Section~\ref{sec:controlled} we test our scores on a dataset where different forms and level of leakage are done in a controlled way so that the ground truth is known
(Section~\ref{genetic_data}); we test the methods on high-dimensional data and study the effect of the choice of the embedding (Section~\ref{main_sec_image_data}). Finally, in Section~\ref{sec:underfit_regime} we observe on a real training example on images that proximity leakage can be detected in the underfitting regime.}

\section{The method} \label{sec:method}
\cf{\subsection{Empirical observation of distinctive NN distributions behavior across generative models}
Our starting point is the observation, made across various instances of datasets and generative models, that the behavior of the NN distance distribution between synthetic and training data can be turned into a stringent criterion for the quality of generated data, as exemplified in Figure~\ref{fig:leak_CDF}. On both panels a generative model is trained on a dataset, and sampled every few epochs. Empirical cumulative distribution functions (eCDFs) of nearest-neighbor (NN) distances allow for distinction of generative scenarios. \textcolor{black}{$d_{TrTr}^*$}, \textcolor{black}{$d_{STr}^*$}, denote the distributions of NN distances from train to train, and synthetic to train respectively. Later on we will also consider \textcolor{black}{$d_{STe}^*$} denoting synthetic to test NN  distribution when extending our proximity based indices from memorization to privacy leakage detection.   We plot the eCDF for each of these distributions and fit the tail of \textcolor{black}{$d_{TrTr}^*$} with an extreme value distribution either Weibull or Gumbel. The x- and y-axes represent the distances and the eCDF, respectively, and are both shown on a $\log_{10}$ scale. When eCDF of \textcolor{black}{$d_{STr}^*$} and \textcolor{black}{$d_{STe}^*$} are below eCDF of \textcolor{black}{$d_{TrTr}^*$}, ie synthetic to train and synthetic to test NN distances are higher than train to train NN distances, we are in the underfitting case. When all eCDFs are aligned, synthetic are indistinguishable from training and test samples. When eCDF of \textcolor{black}{$d_{STr}^*$} is inflated with respect to \textcolor{black}{$d_{TrTr}^*$} and \textcolor{black}{$d_{STe}^*$}, ie distances in \textcolor{black}{$d_{STr}^*$} are lower, then we are in the overfitting case. Proximity-based privacy leaks are detected by measuring the deviation between the \textcolor{black}{$d_{STr}^*$} and \textcolor{black}{$d_{STe}^*$} which may occur both in the overfitting and underfitting regime. The fluctuations of \textcolor{black}{$d_{TrTr}^*$}
around its fit at the bottom of the tail gives the magnitude of the normal fluctuations.}

\subsection{Basic assumptions and hypothesis on nearest-neighbor distances}\label{sec:NNd}
The working hypothesis is that overfitting and privacy leaks that manifest as proximity-based anomalies are reflected in distortions of nearest-neighbor distances between synthetic and train or test samples. \cf{Admittedly this is a restrictive assumption on privacy leakage, which as we will argue can be reached in some cases by an appropriate choice of distances, but which is by no means a general one as many forms of leakage can occur which cannot be detected by proximity metrics}. Then 
the main point underlying the method is that assessing whether a generated sample is too close from a train sample is a statistical question which can be dealt with \textit{extreme value statistics}. 
\cf{The idea is that some statistically meaningful proximity-metrics can be extracted from the behavior of nearest-neighbor distance distribution, as sketch on Figure~\ref{fig:n_excess}. We expect two advantages from that, (compared for instance to the more naive Authenticity score), first by taking into account statistically the influence of dataset sizes. When the size of the synthetic dataset increases the probability to have samples very close to train data increases in a way which is precisely described by EVT. Second, exploiting the fact that NN corresponds to an extreme event regime,  allows us to base  our statistic estimation on a quite robust fit as will be seen experimentally.}

Assume we have a collection $\T_N\! =\! \{\x_i,i=1,\ldots N\}$ of $N$ 
examples, drawn independently from a probability distribution with density $\rho$.
The points $\x_i$ are embedded into a feature space which can be discrete or continuous. In addition we assume that we are given a meaningful distance $d(\x,\y)$ on this space. For genetic sequences for instance it makes sense to use the Hamming distance. 

Given a point $\x$ sampled from the density $\rho$,  
let $\delta(\x,N)$ represent the \textit{nearest-neighbor} (NN) distance of $\x$ to the set $\T_N$ of examples, all sampled as well from $\rho$ :\vspace{-0.2cm}
\[
\textstyle \forall \x\sim\rho, \quad \delta(\x,N) \egaldef d(\x,\y)
\]
where 
\[
\y = \argmin_{\y'\in \T_N}\Bigl[d(\x,\y')\Bigr].
\vspace{-0.2cm}
\]
Then we denote by 
\[
F(u,N) = {\mathbb E}_{\x\sim\rho} \big[ P\left(\delta(\x,N) < u\right) \big]
\]
the probability that this random variable is at a distance smaller than $u$.
For large $N$, we expect the distribution to exhibit a universal behavior in the small-$u$ regime in the form of an extreme value distribution (EVD)~\citep{EVT}. To express it, let us consider
the pairwise distance distribution
between pairs of examples sampled from $\rho$ : \vspace{-0.2cm}
\[
F(u) = \int d^D\x\, d^D\y\, \rho(\x)\rho(\y)\ind{d(\x,\y) < u} \;
\vspace{-0.2cm}
\]

where $d^D$ refers to the infinitesimal volume element in $D$‑dimensional space. From the fact that the NN of $\x$ in $\T_N$ is the result of $N$ independent draws of $\y\sim\rho$ , we have
\[
F(u,N) \leqslant 1-\bigl[1- F(u)\bigr]^N,
\]
where the inequality  becomes an equality when the density $\rho$ is uniform. This upper bound becomes a universal limiting distribution for large $N$, determined by the asymptotic behavior of $F(u)$ as $u \to 0$. In turn this will be dominated by regions of large density for small $u$, which yields a lower bound for $F(u,N)$,  so the lower tail of $F(u,N)$ is expected to coincide with the tail of its upper bound, which justifies to use an EVD fit (more details in supplementary material).

\subsection{Extreme value distribution for nearest-neighbor distances}\label{sec:EVDNN}
When considering NN distances we probe the tail of the pairwise distances distribution.
In that case  we are in the regime of extreme value statistics which displays some universal behavior. 
Extreme value theory (see e.g. ~\citet{EVT}) asserts under some conditions the existence of two sequences $a_N$ and $b_N$ s.t. the rescaled variable $v = (u+b_{N})/a_{N}$
has a limit probability distribution when $N\to\infty$, i.e. 
\[
\lim_{N\to\infty} \bigl[1-F(a_{N}\, v -b_{N})\bigr]^N = G(v),
\]
with $G$ a non-degenerate distribution function representing the probability that the rescaled minimum 
distance $[\delta(\x,N)+b_{N}]/a_{N}$ is greater than $v$. The shift $b_{N}$ tells how $u$ should scale with $N$ while $a_N$ controls the level of fluctuations near this trend. Depending on the tail (either power law or Gaussian) $G$ corresponds to a Weibull or Gumbel distribution (see Supp.Mat. for details).
In practice the fit is performed for a given $N$ with maximum likelihood estimation both on the Weibull and Gumbel cases, and we select the one with highest likelihood. In both cases we have two parameters to determine:
\begin{equation}\label{eq:fit}
  \hat G(u) = \begin{cases}
e^{-\hat A\,u^\alpha}\qquad \text{(Weibull)}, \\[0.2cm]
e^{-\hat A\,e^{\hat B\, u}} \;\;\;\;\, \text{(Gumbel),}
\vspace{-0.1cm}
  \end{cases}
\end{equation}
where $u$ is the non-rescaled distance, the rescaling being implicitly contained in $\hat A$ and $\hat B$, keeping in mind that these fitting coefficients depend on the size $N$ of the dataset against which the NN search is defined and have to be properly  rescaled when e.g. using train and test sets of different sizes. Quality of the fits to the NN distributions was assessed via goodness-of-fit tests and showed close agreement with the EVT hypothesis across datasets and modalities (cf. Section~\ref{sec:GoF_analysis}).

Consider now the situation where we generate $M$ synthetic examples $\{\x_i,i=1,\ldots M\} = \Sc_M$. To each $\x$ of these we associate a NN distance $\delta(\x,N)$ to the set $\T_N$, which is now drawn from the extreme value probability distribution $G$. 
\textcolor{black}{In principle, the train-train NN distances are not independent because $\T_N$ is fixed, therefore, we expect a residual dependency induced by the so-called triangle property of random geometric graphs (~\citet{Penrose2003-cy}). We leave this aside and assume independence as we expect these dependencies to have sub-dominant effects on the analysis (cf Appendix.~\ref{app:dependency_penrose}).}


Given the fit $\hat G$, the probability to find among $M$ NN distances the $r$th smallest one noted $\delta_r(N,M)$  at a value smaller than $u$ is a binomial estimated from the fit as:
\[
P\bigl[\delta_r(N,M) < u\bigr]
 \;\simeq\; \\
 1 - \sum_{q=0}^{r-1}\binom{M}{q}
 \hat G(u)^q\bigl[1-\hat G(u)\bigr]^{M-q}. \vspace{-0.1cm}
\]

Let us call this key quantity $P_{N,M}(u,r)$. This $r$th order statistics represents the probability to actually observe such rank $r$ event whenever the synthetic data are undistinguishable from the real ones. \cf{Table~\ref{tab:assumptions} summarized the key
assumptions and statements behind the estimation of this quantity. The main point being that its relative error is of order $1/\sqrt{N}$ and independent of $M$ under these assumptions.}

\subsection{Scores based on $r$th order statistics}\label{sec:scores}
At this point we have everything needed to compute various overfitting and privacy metrics of interest. \cf{These correspond to various ways of quantifying what we see on nn tail plots like 
the one shown on Figure~\ref{fig:leak_CDF}. For the moment we present them by following some logical way of constructing them and give at the end some guideline to use them.}
First we introduce an individual overfitting score $\pi_r^{\rm ref}$
associated to the sample $\x_r$ extracted from a dataset of interest, relatively to a reference dataset, typically the train or test set, indexed by increasing order w.r.t. NN distances. This score is given by:
\[
\pi_r^{\rm ref} = P_{N,M}(u_r,r)
\]
where $N$ is the size of the reference dataset, $M$ the size of the dataset of interest, 
$u_r$ the $r$-th smallest NN distance to the reference dataset, among the $M$ values computed from the points of the dataset of interest. 
A low value of this score 
indicates that $u_r$ is too small to be at the rank $r$ due to overfitting. As proximity-based privacy leaks may occur also in the underfitting regime~\citet{Yeom_2018}, to detect them we need to take a reference probability based on the test set. This leads us to    
introduce the individual proximity leak score:\vspace{-0.1cm}
\begin{equation}\label{eq:priv_index}
\Delta\pi_r \;\egaldef\; \log_{\rm 10}\Bigl(\frac{\pi_r^{\rm train}}{\pi_r^{\rm test}}\Bigr)\;.\vspace{-0.1cm}
\end{equation}
In other words, for a given synthetic sample, we compute the log ratio of overfitting scores with respect to the training and test sets of real data. A low value indicates a statistical anomaly and suggests a potential proximity-based privacy leak. Unlike standard overfitting, a low privacy score can arise even if the training probability $\pi_i^{\rm train}$ is not particularly small. This leads naturally to the definition of two global proximity leakage indices.
\vspace{-0.1cm}
\begin{equation}\label{eq:global-scores-inline}
\Delta\pi = \frac{1}{M}\sum_{r=1}^M \Delta\pi_r,
\qquad
{\rm NPL} = \sum_{r=1}^M \ind{\Delta\pi_r < \tau}.\vspace{-0.2cm}
\end{equation}

\begin{figure}[ht]
\centerline{\resizebox{0.6\textwidth}{!}{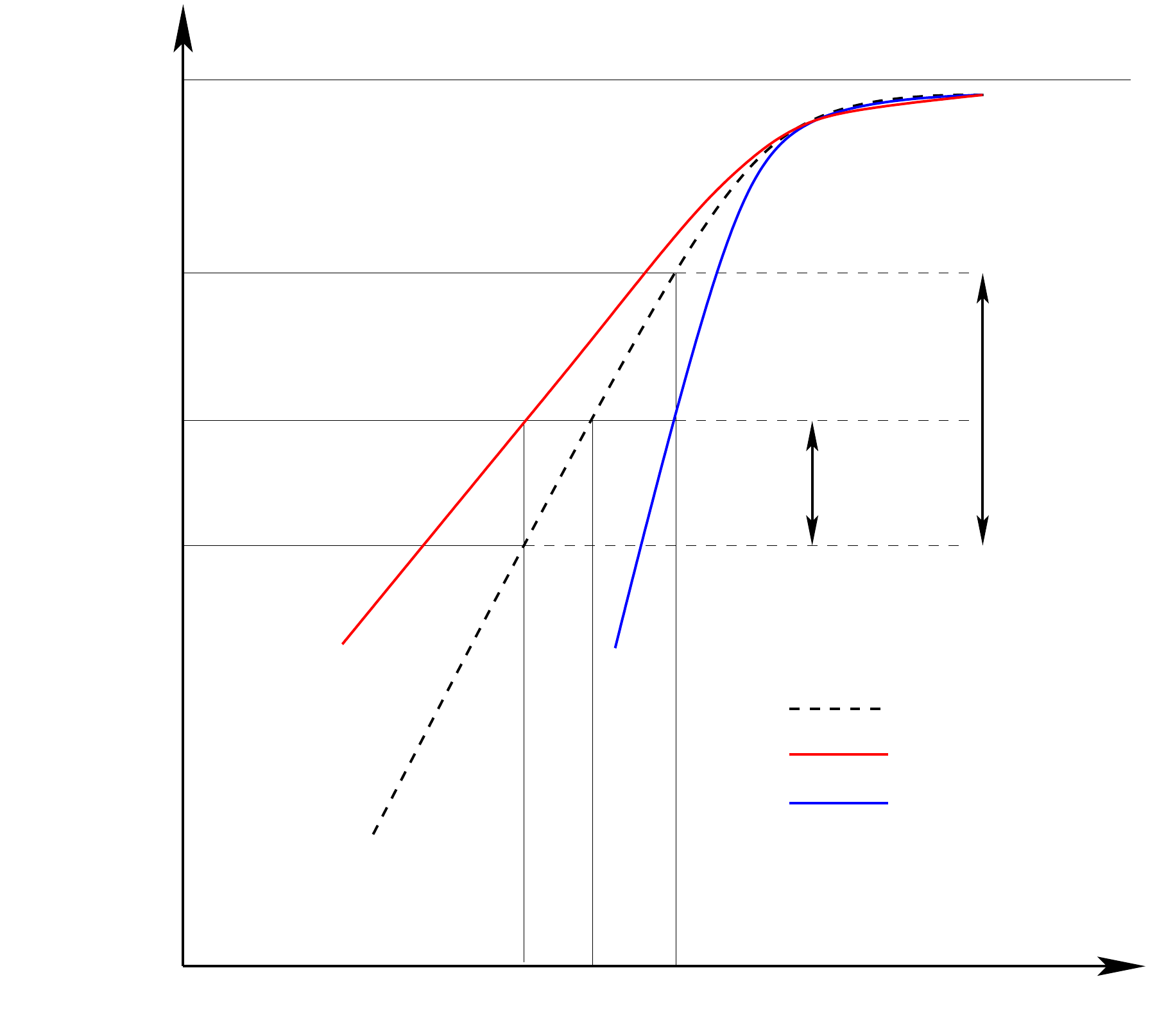}}
\caption{\label{fig:n_excess}\cf{\textbf{Sketch of the method}: possible  scenario where the Synthetic-Train 
[resp. Synthetic-Test] nn distribution stands on the left [resp. on the right] of the reference distribution (fit of the Train-Train), indicating that Synthetic to Train distances are too small (overfitting) while Synthetic to test are too large (underfitting as seen from the test). {\bf $n_{\rm overfit}(r)$ represents for a given rank $r$ the expected number of samples below that rank which are too close to train samples}. $n_{\rm pleaks}$ measures the same thing but relatively to the Synthetic-Test distribution instead of the Train-train, hence representing the same quantity but after {\bf removing the effect of underfitting}}}
\end{figure}

The first one indicates the average level of proximity-based privacy leakage, the second one being the estimated number of synthetic samples breaking privacy given the threshold $\tau$. Typically we will
set $\tau=-3$, corresponding to thresholding a probability of approximately $10^{-3}$, which refers to a rare event.
A more refined quantitative estimation of the number of overfiting samples or proximity-based privacy leaks can be obtained as follows \cf{(see Figure~\ref{fig:n_excess})}. For a given distance $u$ we can obtain this by comparing the actual rank observed for this distance with the rank $r(u)$ corresponding to 
\vspace{-0.2cm}
\[
P_{N,M}(u,r(u)) = \frac{1}{2},
\]
the difference between the two giving the excess (or deficit if negative) of samples having NN distance below $u$. For large $M$ this rank  is directly read off from $\hat G$:\vspace{-0.1cm}
\[
r(u) \approx M\hat G(u) \qquad (r(u) \gg 1) \vspace{-0.1cm}
\]
The reason is that the sum of binomial terms appearing in the definition of $P_{N,M}(u,r)$: 
\vspace{-2mm}
\[
P_{N,M}(u,r) \egaldef 1-\sum_{q=0}^{r-1} \frac{M!}{q!(M-q)!}G(u)^q \bigl[1-G(u)\bigr]^{M-q}
\vspace{-1mm}
\]
when non-vanishing, becomes concentrated on a few terms when $M\gg 1$, the maximum being obtained from the  asymptotics
\vspace{-2mm}
\[
\frac{M!}{q!(M-q)!}G(u)^q \bigl[1-G(u)\bigr]^{M-q} \approx 
\exp\Bigl(-M\bigl[\nu_q\log\frac{\nu_q}{G(u)}+(1-\nu_q)\log\frac{1-\nu_q}{1-G(u)}\bigr]\Bigr)
\]
with $\nu_q = q/M$. Since these terms are symmetric under the exchange $q \leftrightarrow M-q$ the maximum term obtained for $\nu_r \approx G(u) $ corresponds to the rank $r$ s.t. $P_{N,M}(u,r) = \frac{1}{2}$.
As a result $r_u-r(u)$ represents the excess of samples having NN distance below $u$ if $r_u$ is the rank of the sample having NN distance $u$. \cf{More precisely this is the number of samples which should be removed and replaced by other samples having nn distances larger the $u$ in order to have
the distribution coinciding with the reference one at $u$.}
\newcommand{\yes}{\checkmark}
\newcommand{\no}{\ding{55}}

\begin{table}[ht]
\centering
\caption{%
  \textbf{PRIVET scores and their use cases.}
  All scores are built on $\pi_r^{\mathrm{ref}} = P_{N,M}(u_r, r)$,
  the probability of observing the $r$-th ranked synthetic-to-reference
  nearest-neighbor distance under no leakage.
  $M$: number of synthetic samples; $N$: reference set size.
}
\label{tab:privet_scores}
\renewcommand{\arraystretch}{1.55}
\begin{tabular}{
  >{\arraybackslash}p{3.2cm}   
  >{\centering\arraybackslash}p{1.5cm}   
  >{\centering\arraybackslash}p{1.6cm}   
  >{\centering\arraybackslash}p{1.6cm}   
  >{\centering\arraybackslash}p{2.0cm}   
}
\toprule
\textbf{Score} &
\textbf{Granu-larity} &
\textbf{Test set needed} &
\textbf{Threshold needed} &
\textbf{Regime} \\

\midrule

$p_r^{\mathrm{train}}$ \newline
$= \pi_{r-n_{\mathrm{excess}(r-1)}}^{\mathrm{train}}$
& per-sample & \no & \no & memorisation \\

$\Delta p_r$ \newline 
$= \log_{10}\!\bigl( \frac{p_r^{\mathrm{train}}}{p_{r'}^{\mathrm{test}}}\bigr)$ &
per-sample & \yes & \no & overfitting \\

$N_{\rm pleaks}$\newline
$= \max_r n_{\rm pleaks}(r)$
& global & \yes & \no & under-/over-fitting \\

\bottomrule
\end{tabular}
\end{table}

This leads us to define the following quantities:
\textcolor{black}{
\begin{align}
n_{\rm overfit}(r) &\egaldef r - r(u_r^{\rm train}) \label{eq:new_overfit}\\[0.2cm]
n_{\rm pleaks}(r) &\egaldef \min[r,r(u_r^{\rm test}) - r(u_r^{\rm train})] 
\label{eq:new_pleaks}
\end{align}}
where $u_r^{\rm train}$ and $u_r^{\rm test}$ represents  the $r$th
NN distance respectively to the train and to the test dataset.
\cf{The clamp encodes the constraint that among the
$r$ closest synthetic samples there cannot be more than $r$ leaks.
$n_{\rm overfit}(r)$ provides us directly with an estimation, among the first $r$ synthetic samples of lowest NN distance to the train,  of the expected number  of overfitting samples; $n_{\rm pleaks}(r)$, by making use of the NN distance corresponding to the same rank $r$, w.r.t. the test set, actually measures
the excess of small NN distances to the train, compared to the test, among the same first samples $r$. A global proximity-based leakage score can then be straightforwardly obtained from $n_{\mathrm{pleaks}}(r)$, by taking the maximum over the rank
\begin{equation}\label{eq:global_npleaks}
N_{\rm pleaks} = \max_r n_{\rm pleaks}(r)
\end{equation}
From the quantities~(\ref{eq:new_overfit},\ref{eq:new_pleaks}) we can then obtain individual probability scores as follows.} 
First we denote by $n_{\rm excess}(r) \egaldef r-r[u_r^{\rm ref}]$ the expected excess of samples up to rank $r$ whatever reference set is considered (either train of test). Then by correcting from the rank $r$ the excess $n_{\rm excess}(r-1)$ we get an estimation of the individual influence of the sample $r$ 
to the order $r-n_{\rm excess}(r-1)$th statistics which we turn into the probability score
\[
p_r^{\rm ref} = \pi_{r-n_{\rm excess}(r-1)}^{\rm ref}
\]
\cf{The precise meaning of this correction is as follows. The quantity $n_{\rm excess}(r-1)$ represents the number of samples that should be removed up to rank $r-1$ and replaced by samples whose nn distance exceeds $u_r$, in order to eliminate any statistical anomaly in the nn distribution up to rank $r-1$. As a consequence, $p_r^{\rm ref}$ now represents the probability of observing a distance not larger than $u_r$ at rank $r-n_{\rm excess}(r-1)$, under the hypothesis that no proximity anomaly occurs below that rank. In this way, we isolate the contribution of the rank $r$ example alone to the statistical anomaly, if any.
}
From  this we define the individual overfitting score as $p_r^{\rm train}$ and the individual proximity leak score 
\begin{equation}
   \Delta p_r \egaldef \log\frac{p_r^{\rm train}}{p_{r'}^{\rm test}}
   \label{eq:new_score}
\end{equation}
\cf{where $r'$ is the rank  of $\x_r$ in the NN distribution w.r.t. the test set. This has the effect to set a reference $p_{r'}^{\rm test}$ specific to the considered sample in this log ratio.}

\cf{The possible scores that we have defined are summarized in table~\ref{tab:privet_scores}.
The main score is $\Delta p_r$, which is local but usable mainly in the overfitting regime, since the two probabilities involved in the log-ratio rapidly saturate to one in the underfitting regime, making the ratio itself meaningless.
For the underfitting regime we have instead designed $NP_{\rm leaks}$, a global score that could also work in the overfitting regime.
It could possibly be defined as a local indicator (by simply taking a finite difference with respect to the upper rank over which the max is taken), but we have not tested this.
Finally, when no test set is available, it is still possible to use $p_r^{\rm train}$ to check for memorization. The complete procedure used 
to compute these scores is described in~\ref{alg:privet}.
}

\clearpage


\begin{algorithm2e}
\SetKwInOut{Output}{Output}
\SetKwInOut{Input}{Input}
\caption{\textcolor{black}{PRIVET}}\label{alg:privet}
\Input{$Tr \in \mathbb{R}^{N_{Tr}\times D}$ (Train), $Te \in \mathbb{R}^{N_{Te}\times D}$ (Test), $S \in \mathbb{R}^{N_{S}\times D}$ (Synthetic), fit window $[q_1,q_2]$}
\Output{per-sample $\Delta p_r$, per-dataset $N_{\rm pleaks}$}

1) Compute 1-NN distances:\quad $d_{TrTr}^*, d_{STr}^*, d_{STe}^*$\;

2) Fit Weibull and Gumbel to $d^*_{TrTr}$ on window $[\lceil q_1 N_{Tr}\rceil,\,\lceil q_2 N_{Tr} \rceil]$ i.e. find estimators $\hat A, \hat \alpha$ or $\hat A,\hat B$ in~(\ref{eq:fit}) for $\hat G_{d_{TrTr}^*}$.\;

3) \If{$N_{Tr} \neq N_{Te}$}{
$\lambda\leftarrow N_{Tr}/N_{Te}$ \;
\If{Weibull}{$\kappa\leftarrow\lambda^{-1/\hat\alpha}$ \;$d^*_{STe}\leftarrow\kappa\,d^*_{STe}$}
\If{Gumbel}{$d^*_{STe}\leftarrow d^*_{STe} - \frac{\ln(\lambda)}{\hat B}$}  
}

4) Sort $d^*_{STr}$ ascending: $\delta_1^{STr}\!\le\!\dots\!\le\!\delta_{N_S}^{STr}$,
rank map $j\mapsto r(j)$, abbr. as $r$\;
Sort $d^*_{STe}$ ascending: $\delta_1^{STe}\!\le\!\dots\!\le\!\delta_{N_S}^{STe}$,
rank map $j\mapsto r'(j)$, abbr. as $r'$\;

5) Write $\pi_r^{ref} = 1 - \sum_{q=0}^{r} \binom{N_S}{q}
  \left[\hat G_{d_{TrTr}^*}(\delta_{r}^{S\text{-to-}ref})\right]^q
  \left[1 - \hat G_{d_{TrTr}^*}(\delta_r^{S\text{-to-}ref})\right]^{N_S - q}$\;

\For{$r \in \{1,\dots,N_S\}$}{
  $n_{\rm pleaks}(r)\leftarrow
     \min (r, (\lfloor N_S \hat G(\delta_r^{STe}) \rfloor-\lfloor N_S \hat G(\delta_r^{STr}) \rfloor)))$\;

  $p_r^{\rm Tr} \leftarrow \pi_{r-n_{\rm excess}(r-1)}^{\rm Tr}$;  $p_{r'}^{\rm Te} \leftarrow \pi_{r'-n_{\rm excess}(r'-1)}^{\rm Te}$; $\Delta p_r \leftarrow \log\frac{p_r^{\rm train}}{p_{r'}^{\rm test}}$


}
\Return $\{\Delta p_r\}_{r \in \{1,\dots,N_S\}}$, $N_{\rm pleaks}$\;
\end{algorithm2e}

\section{Assessing the quality of fits to the NN distributions}
\label{sec:GoF_analysis}

\begin{figure*}[!h]
\centering
\includegraphics[width=1\textwidth]{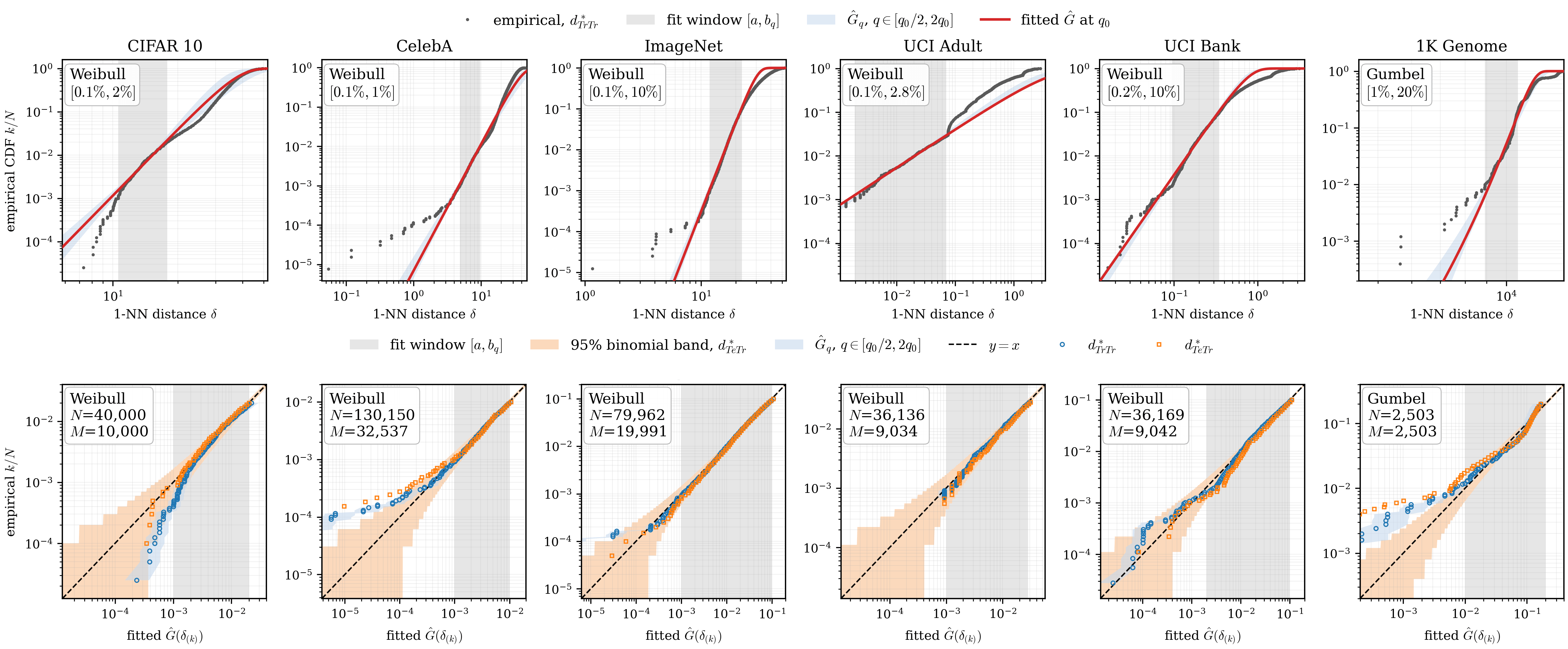}
\caption{\textcolor{black}{\textbf{Nearest-neighbour distances follow an extreme value tail law across datasets and modalities.} \textbf{Top.} eCDF of the train-to-train 1-NN distances (black) with fitted law (red) on the corresponding window (grey). The blue ribbon repeats the estimation for $q_2\in[q_2^*/2,2q_2^*]$. \textbf{Bottom.} Each point is one order statistic: on the x-axis is the probability the fitted law assigns to that distance,
$\hat G(\delta_{(k)})$, and on the y-axis is the empirical cumulative, $k/N$. Circles are train-to-train distances
($d^*_{TrTr}$), squares are test-to-train distances ($d^*_{TeTr}$), grey band is the fit window. The blue ribbon repeats the estimation for $q_2\in[q_2^*/2,2q_2^*]$, showing stability to the upper end of the fit window hyperparameter, with $q_2^*$ being the initial hyperparameter choice. The orange region is the binomial confidence band at a 95\% confidence level for the test-train distances and answers the following question:  if the model's predicted probability $p$ were true, and $M$ independent held-out points were drawn, their observed fraction below the corresponding distance would land between the $2.5$th and $97.5$th percentiles band 95\% of the time. $N$ and $M$ denote the training and test set sizes, respectively.}}
\label{fig:GoF_main}
\end{figure*}

\textcolor{black}{We perform a meta-analysis across datasets and modalities to empirically assess
the extreme value assumption. For each dataset we estimate the tail law on a
window $[q_1,q_2]$ of the train-to-train 1NN distances and compare the probability assigned by the fitted law, with the corresponding empirical cumulative probability. The comparison is carried out both on the train-to-train distances, from which the law is estimated, and on the test-to-train distances, which the fit never sees and which therefore provide an independent diagnostic. The window hyperparameter $[q_1,q_2]$ is selected per dataset by inspecting the eCDF and placing it inside the first mode when multimodality is observed (Fig. Top.~\ref{fig:GoF_main}). Indeed, the EVT approximation concerns the lower tail rather than the full NN distribution, the fitting interval is thus chosen  within the low-distance regime. We also conduct goodness-of-fit diagnostics to evaluate the stability of the fits with respect to this hyperparameter choice. More details about the procedure are given in Appendix.~\ref{sec:GoF_analysis}.}

\textcolor{black}{In Fig. Bottom.~\ref{fig:GoF_main}, we first observe that inside the window the train-to-train empirical distribution vs fitted law lie on the diagonal for every dataset, thus showing that the fit describes the data where it is intended to. This agreement is also observed on the held-out test-to-train distances, which are not used for the EVD estimation. The blue ribbon, which repeats the estimation for eleven values of $q_2$ ranging from half to twice the selected one, is narrow throughout, indicating that the fit has little sensitivity to the window in the tested range. Below the fitting window, where the law is extrapolated, deviations from the diagonal become visible for some datasets at the smallest probabilities. Their interpretation is limited by the decreasing number of held-out observations available in this regime, as reflected by the widening of the binomial band. The largest deviations are observed for CelebA and 1000 Genomes data. In Section. \ref{genetic_data}, PRIVET is additionally successfully assessed on controlled genetic data experiments that directly evaluate the practical consequences on the downstream scores. We also include the per-sample score distribution of $\Delta p_r$ under a no-leakage null on CIFAR-10 and 1000 Genome data (FIG.~\ref{fig:noleakage_null_score_distrib}).}



\textcolor{black}{We further assess whether the fitted tail law depends on the particular partition of the data by repeating the estimation on $R=10$ independent random splits (Appendix.Fig.~\ref{fig:GoF_splits}). Each replicate shuffles the dataset with its own seed, divides it $80/20$, recomputes the train-to-train nearest-neighbour distances on the larger part, fits a new extreme-value law on that dataset's window, and evaluates the frozen law on the held-out part, which took no part in the fit.  The fitted curves remain closely grouped across partitions and their median generally follows the diagonal, indicating that the inferred tail is reproducible with respect to the random data split. The variability increases towards the smallest probabilities, where fewer observations are available, and is largest for CelebA and the 1000 Genomes data. This diagnostic therefore complements the goodness-of-fit analysis by assessing the sensitivity of the EVD tail-estimation procedure to the particular subset of observations used for fitting.}

Overall, this meta-analysis supports the conclusion that the asymptotic behavior of distances follows the extreme value regime across all examined datasets and the EVT tail assumption is a general one.


\section{Controlled experiments}  \label{sec:controlled}
\subsection{Genetic data: \textcolor{black}{PRIVET is sensitive to leakage}}
\label{genetic_data}

Here we perform a controlled experiment where a certain amount of information from the training data is leaked into the pseudo-synthetic data. 
These data are real genetic data corresponding to binary sequences of size $65535$, retrieved from \citet{the1000genomesprojectconsortiumGlobalReferenceHuman2015} (Appendix~\ref{1KG_data} for further details). Distance between samples is the Hamming distance.
We partition these $5004$ samples into three random subsets each of equal size $1668$ and assign these to train, test and synthetic datasets, the latter corresponding to the output from a
fictional generative model. Then we set two parameters, $f_{\rm fake}\!\in\![0,1]$ and $f_{\rm copy}\!\in\![0,1]$, corresponding respectively to the fraction of
synthetic samples containing leaked information from the training data (i.e., percentage of copied samples)  and the
fraction of leaked information within such a sample (i.e., percentage of copied segments). The synthetic data are constructed as follows. In the subset corresponding to synthetic data, a random fraction $f_{\rm fake}$ is selected at random to contain leaked information. For each of these assigned samples, we first identify its $k$-nearest neighbor in the training set and replace in its sequence a fraction
$f_{\rm copy}$ of random bits copied from the training sample, which model the private information assumed to be leaked by the generative model. \textcolor{black}{Since a realistic leak might be more likely to occur within a specific mode (such as populations for genetic data, where a given population has $\sim200$ maximum samples), we take $k \in \{1,2,3,4,5,200\}$ to avoid overestimating performance of 1-NN based methods}. To demonstrate robustness, we also repeated the experiment on a low-sample-size scenario dataset of 198 samples (Appendix~\ref{app:controlled_leak_low_sample}). 

\begin{figure*}[!h]
\centering
\includegraphics[width=1\textwidth]{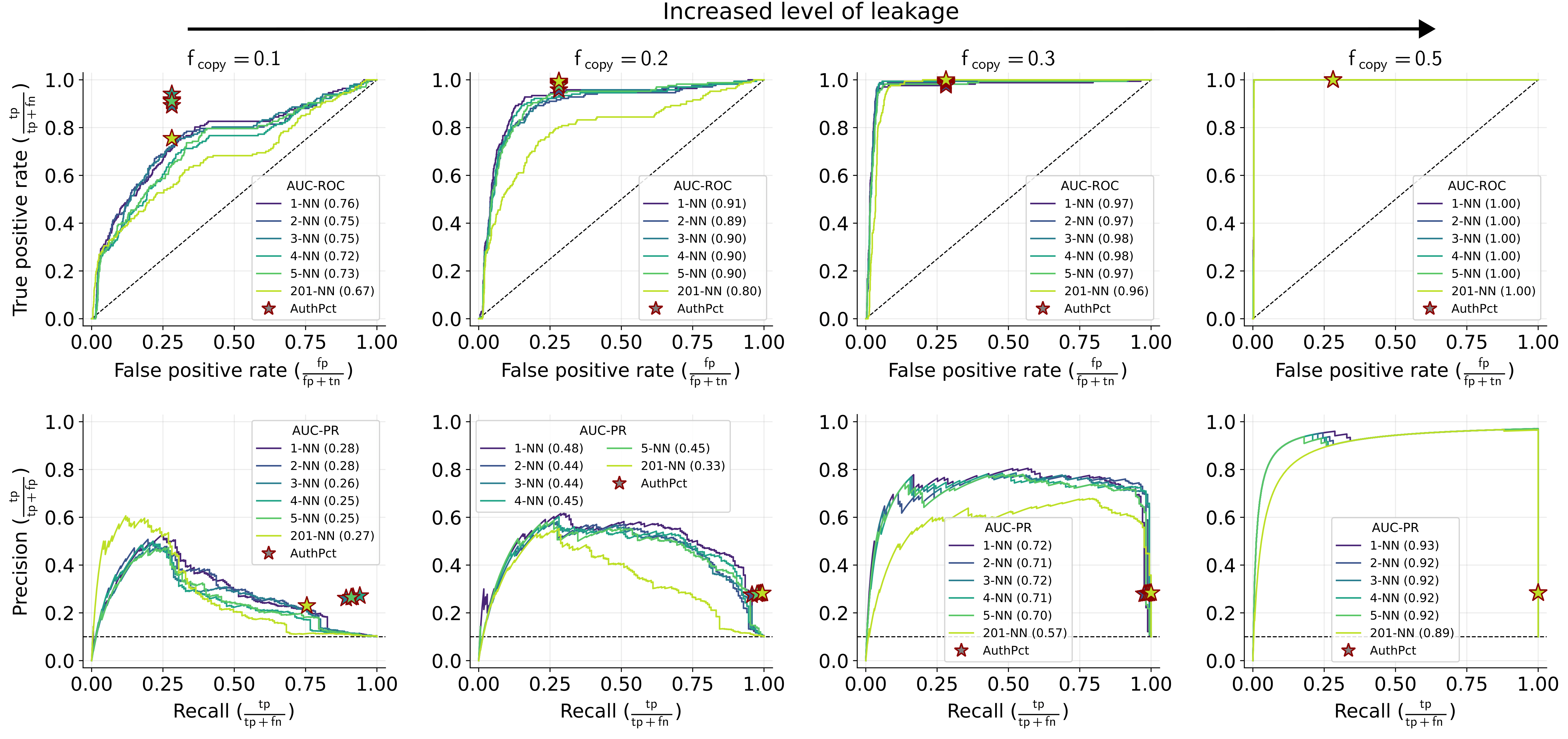}
\caption{\textcolor{black}{\textbf{PRIVET is sensitive to leakage. }}\textcolor{black}{ROC (top row) and Precision-Recall (bottom row) curves for PRIVET's $\Delta p$ score. We set $f_{\mathrm{fake}}=0.1$ and vary $f_{\mathrm{copy}}$. The performances for PRIVET (curves) and Authenticity (red star) are shown for various leakage designs based on $k$-NNs with $k \in \{1,2,3,4,5,200\}$. The black dashed lines corresponds to the random baseline. Authenticity uses a fixed binary decision rule, yielding a single point rather than a curve.}}
\label{fig:genetic_leakage_roc_pr_tmlr}
\end{figure*}

This controlled experiment on genetic data serves as a sanity check for our metric. \textcolor{black}{In Fig.~\ref{fig:genetic_leakage_roc_pr_tmlr}, PRIVET's $\Delta p_r$ score exhibits sensitivity to the $f_{\rm copy}$ leakage mechanism across all $k$-NN regimes, with an accurate identification of leaked samples. On the ROC curves (top row), for  $f_{\rm copy} \geq 0.2$, PRIVET attains comparable  true positive rates to Authenticity while operating at lower false positive rates. On the 
Precision-Recall curves (bottom row), Authenticity achieves high recall at the cost of systematically lower precision than PRIVET.Further experiments in the supplementary material (Appendix,~\cref{fig:privacy_map_65K,fig:privacy_map_glocal_low_sample_size,fig:privacy_map_groundtruth_low_sample_size,fig:privacy_map_groundtruth_65K,fig:local_implicit_decimation,fig:local_decimation})
confirm that PRIVET reliably detects proximity-based privacy leakage under controlled conditions, whereas alternative local or global scores appears noisy or insufficiently sensitive.}

\subsection{Image data: \textcolor{black}{PRIVET is competitive with baselines and highlights the importance of representation quality}}
\label{main_sec_image_data}

PRIVET is a general-purpose algorithm applicable across diverse data modalities, provided the feature representations yield well-behaved distance metrics. To evaluate its versatility, we also benchmarked PRIVET on the copycat experiment ~\citep{copycat_OG,NEURIPS2023_68b13860}. We used synthetic data generated by SOTA diffusion model on CIFAR10 ~\citet{pfgmpp} that was provided by ~\citep{stein}. 

We applied several transformations on training and synthetic data corresponding to easy and hard distortion scenarios, that preserve information while distorting geometry, then created pseudo-synthetic datasets consisting in progressive replacement of transformed synthetic data with transformed training data, and computed embeddings of these data using either DINOv2 ~\citep{oquab2024dinov}. Finally, we applied the metrics on the embeddings of the non-transformed training set, non-transformed test data and the mixture of transformed synthetic and transformed training data (pseudo-synthetics). This corresponds to a situation where the generative model produces distorted copies of the training set. \textcolor{black}{To demonstrate robustness, we also repeated the experiment on a low-sample-size scenario dataset of 100 \& 1000 samples (Appendix. FIG. ~\ref{fig:copycat_dinov2_lowdata} \& \ref{fig:copycat_dinov2_lowdata_grayscale}).}

\begin{figure*}[!h]
\centering
\includegraphics[width=1\textwidth]{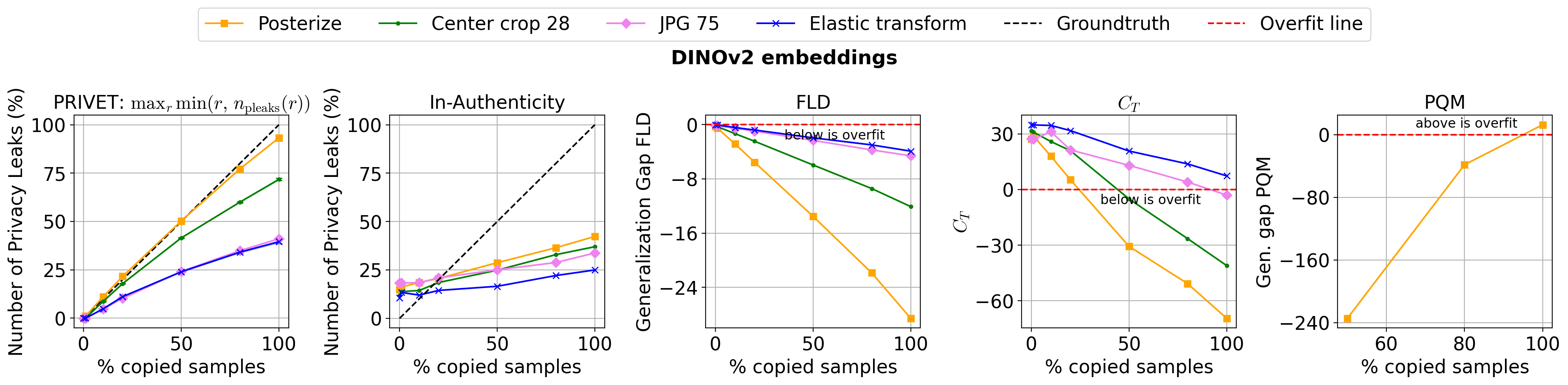}
\caption{\textcolor{black}{\textbf{PRIVET is competitive with baselines and highlights the importance of representation quality}} Several computer vision transformations (posterize, center crop 28, JPG 75, Elastic transform) are applied on CIFAR10 synthetic and training samples to generate a pseudo-synthetic dataset. x-axis represents the fraction of synthetic samples replaced with training ones; y-axis reports the value of the corresponding metric. For PRIVET and In-Authenticity, we report the percentage of identified proximity-based privacy leaks. \textcolor{black}{For PRIVET we report $N_{\rm pleaks}$, shaded bands and error bars give the spread over five random draws of the reference and holdout sets and of the injected samples, at a fixed generator checkpoint. However, these are tiny as there is substantial overlap between draws from different seeds.} For Gen. gap (FLD) and $C_T$, overfitting is detected when the metric falls below the horizontal dashed line, whereas for Gen. gap (PQMass), overfitting is indicated when the metric exceeds the threshold. DINOv2 (ViT-B/14 distilled without registers) embeddings of CIFAR10 were computed. For several transformation PQMass score diverged to $- \infty$, hence results are omitted.}
\label{fig:copycat_dinov2}
\end{figure*}

\textcolor{black}{In the case of PRIVET, we report the leakage count $N_{\rm pleaks}$
given in Eq.~\ref{eq:global_npleaks}.} This choice is motivated by the particular behavior induced by DINOv2 embeddings: although embeddings of PFGM++ samples generally appear underfit, the posterize and center crop 28 transformations produce leakage patterns in which some synthetic samples look overfit while others remain underfit but still exhibit tail deviations (Appendix.FIG.\ref{app:cdf_copycat}A. for CDFs of 1-NN distances). PRIVET reliably detects proximity-based privacy leakage, as half of the curves closely follow the ground truth (Fig. \ref{fig:copycat_dinov2}) and outperform In-Authenticity on all four. \textcolor{black}{Crucially, the estimate at $\beta=0$ is zero for every transformation, so the method does not manufacture leaks in the absence of any.} In contrast, the In-Authenticity metric shows a mild increase with the fraction of copied samples: it begins with false positives and quickly transitions to underestimating the number of true leaks. Finally, both the generalization gap of FLD and the $C_T$ score trend in the expected direction; however, FLD starts in an overfitting regime, while $C_T$ only signals overfitting when a substantial amount of leakage is present. \textcolor{black}{This copycat experiment demonstrates detection through several distorting transformations (posterize, crop, JPEG compression, elastic deformation, grayscale), all of which alter pixel-level distances substantially while remaining detectable in embedding space.}

\textcolor{black}{Replacing DINOv2 with wavelet packet coefficients \citet{veeramacheneni2025fwd} further highlight the impact of the embedding on downstream evaluation metrics (Appendix. Fig.~\ref{fig:copycat_dinov2_app}). While PRIVET is modality-agnostic by construction, its performance naturally reflects the quality of the chosen embedding, and as such can also provide a diagnostic of the representation itself, thus enabling comparison of embedding spaces. In practice, we recommend inspecting the 1-NN CDFs, which provide immediate feedback on the regime (underfitting or overfitting) induced by a given embedding. This visual diagnostic also helps ensure that PRIVET operates in a representation where its distance-based assumptions are valid. In practice, some data types benefit from an embedding space while others do not. The choice of representation is left to the practitioner, who should select the space that best captures the relevant structure of the data.}

\section{\textcolor{black}{PRIVET accurately detects data leakage in the underfitting regime}\label{sec:underfit_regime}}
\textcolor{black}{We further diagnose whether data leakage in the underfitting regime is genuine rather than a distributional artifact. For this purpose, we trained a diffusion model on a subset of $1{,}000$ images from CIFAR-10 for $1{,}000$ epochs and generated $1{,}000$ samples. We then compute the 1NN distances on the DINOv2 embeddings of those images. Further details on training the diffusion model can be found in Appendix.\ref{ddpm_cifar10_1k}.}

\begin{figure*}[!h]
\centering
\includegraphics[width=0.9\textwidth]{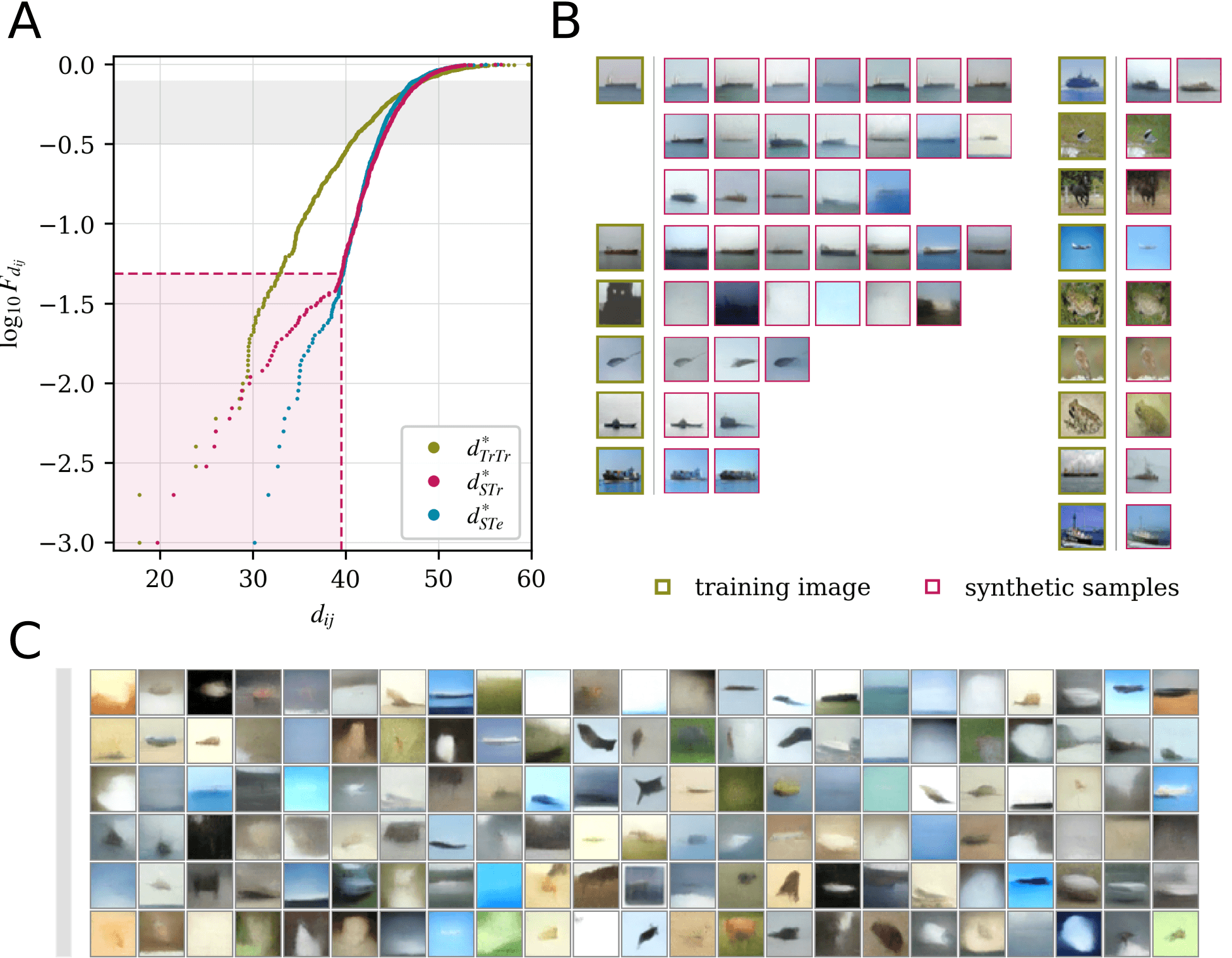}
\caption{\textcolor{black}{\textbf{Data leakage in the underfitting regime is genuine and detectable by PRIVET. A.} Empirical CDFs of train-to-train (olive), synthetic-to-train (magenta), and synthetic-to-test (cyan) 1-NN distances. The red shaded region highlights the tail where the synthetic-train and synthetic-test distributions deviate, whereas the grey shaded region indicates a region in the bulk of $d^*_{STr}$. \textbf{B.} Examples from the deviating tail of synthetic-to-train nearest neighbor pairs. The left column shows training images (olive frames), while the remaining columns show their synthetic counterpart (magenta frames), revealing memorized training examples among synthetics. \textbf{C.} Synthetic samples from the bulk of $d^*_{STr}$, illustrating the predominantly underfit outputs of the generative model.}}
\label{fig:dataleakage_underfitting}
\end{figure*}

\textcolor{black}{
As shown in Fig.~\ref{fig:dataleakage_underfitting}.A, this learned generative model results in a scenario with proximity-based data leakage in the underfitting regime, with pronounced tail deviation between CDFs of synthetic-train and synthetic-test 1NN distances. Inspection of synthetic-training pairs within this deviating tail (Fig.~\ref{fig:dataleakage_underfitting}.B) confirms that the corresponding synthetic samples are memorized training examples. In contrast, synthetic samples drawn from the bulk of $d^*_{STr}$ (Fig.~\ref{fig:dataleakage_underfitting}.C) are predominantly underfit: the model either generates blurry images or captures only coarse, low-frequency features such as background color and overall shape. This qualitative inspection confirms that the observed underfitting is genuine rather than an artifact of the DINOv2 representation, while demonstrating that data leakage can nevertheless occur in this regime.
}

\section{Conclusion}

\begin{tcolorbox}[breakable,
  colback=PastelBlueBg,
  colframe=PastelBlueFrame,
  coltitle=black,
  title={\textbf{Guide for the user}},
  boxrule=0.4mm,
  arc=1mm
]
In practice, we recommend first inspecting the train-to-train 1-NN CDF to assess agreement with the extreme value tail assumption, and then comparing the synthetic-to-train and, when available, synthetic-to-test 1-NN CDFs to identify PRIVET's operating regime: underfitting with or without data leakage, or overfitting.

\begin{center}
\resizebox{0.8\textwidth}{!}{%
\begin{tikzpicture}[
    font=\small,
    q/.style   = {align=center, inner sep=3pt},
    s/.style   = {align=center, inner sep=5pt, draw, line width=.7pt},
    reg/.style = {align=center, font=\footnotesize, text=black!75, inner sep=2pt},
    lbl/.style = {font=\footnotesize, inner sep=1.5pt},
    ->, >=latex, line width=.6pt
]
\node[q] (root) at ( 0.0, 0.4) {what is scored?};
\node[q] (glob) at (-3.2,-0.7) {a dataset};
\node[q] (loc)  at ( 3.2,-0.7) {a sample};
\node[q] (gq) at (-3.2,-1.9) {is a test set available?};
\node[q] (lq) at ( 3.2,-1.9) {is a test set available?};
\node[s] (npl) at (-5.0,-3.5) {$N_{\mathrm{pleaks}}$};
\node[s] (nov) at (-1.6,-3.5) {$N_{\mathrm{overfit}}$};
\node[s] (dp)  at ( 1.6,-3.5) {$\Delta p_r$};
\node[s] (pr)  at ( 5.0,-3.5) {$p_r^{\mathrm{train}}$};
\node[reg] at (-5.0,-4.5) {over- /\\under-fitting};
\node[reg] at (-1.6,-4.5) {overfitting};
\node[reg] at ( 1.6,-4.5) {overfitting};
\node[reg] at ( 5.0,-4.5) {memorisation};
\draw (root) -- (glob);
\draw (root) -- (loc);
\draw (glob) -- (gq);
\draw (loc)  -- (lq);
\draw (gq) -- node[lbl,pos=.55,left]  {yes} (npl);
\draw (gq) -- node[lbl,pos=.55,right] {no}  (nov);
\draw (lq) -- node[lbl,pos=.55,left]  {yes} (dp);
\draw (lq) -- node[lbl,pos=.55,right] {no}  (pr);
\end{tikzpicture}}

\textbf{Decision tree.}
\end{center}
\end{tcolorbox}

In this work, we presented a novel framework based on extreme value theory, PRIVET \footnote{https://anonymous.4open.science/r/PRIVET-28FA}, for detecting memorization, overfitting and proximity-based privacy leakage in underfitting regime, in synthetic datasets. Our method can provide both sample-level and dataset-level quantitative assessment of proximity-based privacy leaks from the training dataset, with the further possibility of assigning a boolean "leak-or-not" flag for each synthetic sample. We tested PRIVET for genetic and image data domains under controlled settings, where the amount of privacy leak is known, and in a realistic setting: an underfitting generation setting producing data leakage. Results on image data demonstrate that PRIVET captures forms of proximity-based privacy leakage arising from transformations that preserve sensitive information while distorting geometry, \textcolor{black}{ and that representation quality bounds what any distance-based metric can detect}. We compared our method to existing ones and reported competitive performance under various settings. \textcolor{black}{The asymptotic behavior of nearest-neighbor distances follows the expected Extreme Value regime not only for images and genetic sequences but also for heterogeneous tabular data, as shown in Fig.~\ref{fig:GoF_main}.}
To the best of our knowledge, our method is the first to offer a comprehensive assessment of both individual samples and entire datasets, while remaining theoretically grounded, scalable, and robust across low-sample regimes and diverse domains. 

\section*{Broader Impact Statement}
The growing use of generative AI for synthetic data generation raises serious privacy concerns, particularly when models are trained on sensitive data such as copyrighted content or human genetic information governed by privacy regulations. Although deep generative modeling has seen rapid progress, the development of rigorous evaluation methods, especially for privacy auditing, has not kept pace. Existing privacy assessments typically rely on global, dataset-level metrics that offer limited interpretability and overlook individual-level risks. In this work, we address this gap by introducing a sample-level privacy metric capable of detecting overfitting and privacy leaks in synthetic datasets. Our method, PRIVET, provides interpretable, fine-grained privacy scores and is applicable across data modalities. We believe this contributes a necessary layer of accountability and transparency to the deployment of generative models, particularly in high-stakes domains like biomedical research and content-sensitive applications.


\bibliography{tmlr}
\bibliographystyle{tmlr}

\newpage
\appendix
\onecolumn

\section{Nearest neighbor distances and EVD}\label{app:NN_EVD}
Let us justify more precisely here the fit which is performed on NN distributions.
Taking the notations of Section.\ref{sec:NNd} we want to characterize the 
distribution of the random variable $\delta(\x,N)$ where $\x\sim\rho$.
Let us call
\begin{align*}
F_{\x}(u) &\egaldef P_{\y\sim\rho}\bigl[d(\x,\y) < u \bigr], \\[0.2cm]
F_{\x}(u,N) &\egaldef P_{\T_N\sim\rho^N}\bigl[\delta(\x,N) < u \bigr] \\
& = P_{\T_N\sim\rho^N}\bigl[ \min_{\y'\in \T_N} d(\x,\y') < u \bigr] \\
& = 1 - P_{\T_N\sim\rho^N}\bigl[ \min_{\y'\in \T_N} d(\x,\y') \geqslant u \bigr] \\
& = 1 - P_{\y_1\sim\rho, ..., \y_N\sim\rho}\bigl[ \forall \y'\in \{\y_i\}_{i=1}^{N},\; d(\x,\y') \geqslant u \bigr] \\
& = 1 - \left( P_{\y\sim\rho}\bigl[ d(\x,\y) \geqslant u \bigr] \right)^N \\
& = 1 - \big( 1 - F_{\x}(u) \big)^N \; .
\end{align*}
We have 
\[
F(u,N) \;=\; 
\EE_{\x\sim\rho} \bigl[ F_{\x}(u,N) \bigr] \;=\;
1- \EE_{\x\sim\rho} \Big[ \bigl(1-F_{\x}(u)\bigr)^N \Big] \; .
\]
Since by definition
\[
F(u) = \EE_{\x\sim\rho} \bigl[F_{\x}(u)\bigr]
\]
by convexity of the function $x\to (1-x)^N$ we get 
\[
F(u,N) \leqslant 1 - \bigl[1-F(u)\bigr]^N
\]
with equality when the function $u \mapsto F_{\x}(u)$ does not depend on $\x$. Hence \textbf{in the case of a uniform distribution over a manifold, this bound is an equality}. Indeed in the limit when the number $N$ of samples tends to infinity, the distribution of distances to nearest neighbors is the same at all locations $\x$: every point $\x$ has the same density and a neighborhood of same dimensionality, which is locally identical in the infinitesimal limit (tangent space). 

In the case of more complex distributions, let us call 
\[
\bar G(v)  \egaldef \lim_{N\to\infty} \bigl[1-F(u)\bigr]^N
\]
with $v$ the appropriate rescaled variable, depending on the universality class of the EVD.
Indeed, depending on the behavior of the lower tail of the distribution $F$, the corresponding extreme value statistics either converges to a Weibull  distribution:
\begin{equation}\label{eq:Weibull}
G(v) = \exp(-A v^\alpha),
\end{equation}
with $a_N = N^{-1/d}$ and $b_N=0$, when $F(u)$ has typically a power law tail with exponent $\alpha$; either to a Gumbel distribution if for instance $F(u)$ has a Gaussian tail again at small $u$:
\begin{equation}\label{eq:Gumbel}
G(v) = \exp\bigl(-e^v\bigr), 
\end{equation}
after centering and normalizing the relative distance variable $u\to \tilde u= (u-\bar u)/\sigma$ w.r.t. 
to mean $\bar u$ and std $\sigma$, with a scaling $v = (\tilde u + b_{N})/a_{N}$ now given by $b_{N} = \sqrt{2\log(N)} +\mathcal{O}\Bigl(\log\log(N)\Bigr) = 1/a_{N}$ see~\cite{EVT}.

So as we see $F(u,N)$ is asymptotically a mixture of EVD, indexed by $\x$, 
which is dominated by the EVD defined by the averaged pairwise distance distribution.
To see how this mixture behaves, first we make the hypothesis of a data manifold of dimension $d$
so that the average over $\x$ and $\y$ to get $F(u)$ is to be understood as
\[
F(u) = \int d\x^d d\y^d \rho(\x)\rho(\y) \ind{d(\x,\y)\le u},
\]
where $\rho$ is to be considered as the probability density of the data on this manifold, and $d(\x,\y)$ a distance induced by a metric on the manifold. The metric is chosen in such a way that the expected number of samples inside a sphere of infinitesimal radius $\epsilon$ centered on $\x$ scales like $\rho(\x)\epsilon^d$, meaning in practice that the length on all directions are measured in comparable units. With these hypothesis we basically have that when sampling  $\y\sim\rho$
the radial probability density w.r.t. some point $\x$ is locally given by the distribution 
\[
f_{\x}(u) = P_{\y\sim\rho}[d(\x,\y) \in du] \propto \rho(\x) du u^{d-1} e^{-V_{\x}(u)}
\]
where $V_{\x}$ is in principle dependent on $\x$, but we assume it to be a smooth potential with $V(u)\to\infty$ at large $u$ and to continuous and differentiable when $u$ is close to zero.

\paragraph{Power law tail:} For $d=\mathcal{O}(1)$, the corresponding 
probability density behave like $F_{\x}(u) \sim \rho(\x) u^d$ at small $u$, hence when $N$ get sufficiently large, then the NN neighbor distance to $\x$ when sampling $N$ samples $\y\sim\rho$ will scale like $u_N\sim[N\rho(\x)]^{-1/d}$, leading us to a Weibull distribution with exponent $\alpha = d$ independent of $\x$ and $a_N\sim [N\rho(\x)]^{-1/d}$.
So, with these hypothesis we expect the coefficient $\hat A$ of the Weibull fit~(\ref{eq:fit}) to depend on $\rho(\x)$ but not the exponent.
\paragraph{Gaussian tail:} Instead when $d\gg 1$, $f_{\x}$ becomes Gaussian around the point $\tilde u$
satisfying $V_{\x}'(\tilde u_{\x})-(d-1)/\tilde u_{\x} = 0$. So we then get a Gaussian tail, leading us to a Gumbel distribution, but in principle both 
$\tilde u_{\x}$ and the standard deviation $\sigma_{\x}$ should depend on $\x$. If this dependency can be neglected, then the coefficient $\hat B$ in the 
Gumbel fit~(\ref{eq:fit}) is independent of $\x$. We will assume that in the following. By contrast, the other coefficient $\hat A$  should depend in any case
on $\rho(\x)$.
So considering now $F(u,N)$ when $N$ is large under the hypothesis of a data manifold we have
\[
\lim_{N\to\infty} F(u,N) =  \int d^d\x \rho(\x)[1-G_{\x}(v)]
\]
If we look at the tail, i.e. $v\to 0$ for Weibull and $v\to -\infty$ for 
Gumbel, we have
\[
\lim_{N\to\infty} F(u,N) \approx 
\begin{cases}
\DD \bar A v^\alpha,\qquad (\text{Weibull case}) \\[0.3cm]
\DD \bar A e^{Bv},\qquad (\text{Gumbel case})
\end{cases}
\]
with
\[
\bar A = \EE_{\x\sim\rho}\Bigl[A_{\x} \Bigr]
\]
In addition, we have the fact that in both cases the coefficient $A_{\x}$ inherits his dependency on $\x$ from $F_{\x}(u)$ which is of the form when $u$
is close to zero
\[
F_{\x}(u) \approx \rho(\x) 
\begin{cases}
\DD u^\alpha,\qquad (\text{power law tail}) \\[0.2cm]
\DD \int_{-\infty}^u dr e^{-\frac{(r-\tilde u)^2}{\sigma^2}},\qquad (\text{Gaussian tail})
\end{cases}
\]
As a result, the average which is done beforehand in $\bar G$ leads to the same
constant fit  $\bar A$ which somewhat justifies (given the necessary hypothesis) the fit we are doing.

\section{Various scores based on order $r$th statistics} \label{sec:various-scores}
$\pi_r^{\rm ref} = P_{N,M}(u,r)$ is the key statistic that we use to detect overfitting and privacy leaks. In Section~\ref{sec:scores} we propose one straightforward individual privacy score
\[
\Delta\pi_r = \log\frac{\pi_r^{\rm train}}{\pi_r^{\rm test}}
\]
which yields good results, but other 
possible definitions exist. The rational behind this log ratio is to condition the order $r$th
statistics for the train on the $r$th statistics for the test
\[
\frac{\pi_r^{\rm train}}{\pi_r^{\rm test}} = P(\delta_r(N,M) < u_r^{\rm train} \vert \delta_r(N,M) < u_r^{\rm test}),
\]
where $\delta_r(N,M)$ is the random variables corresponding to the ranked $r$th NN distance among $M$ synthetic data from a reference set of size $N$, following $G(u)$, while $u_r^{\rm train,test}$
are the value of these distances actually observed respectively w.r.t. the train and to the test set.
This obviously make sense when $u_r^{\rm train} < u_r^{\rm test}$ which is most generally observed.

A very small value of $\pi_r^{\rm ref}$ is telling us that the $r$th NN distance $u_r$ is too small, or stated differently 
there are too many samples with NN distance below $u_r$. In contrary a very small value of $1-\pi_r^{\rm ref}$ that there are too few samples with NN distance below $u_r$. 
$\Delta\pi_r$ is therefore meaningful mostly in the over-parameterized
regime, because in the under-parameterized regime both  probabilities $\pi_r^{\rm train,test}$
rapidly saturate to one when the tail of the NN distribution of $d^*_{STr}$ and $d^*_{STe}$ are well below the expected one obtained from $d^*_{TrTr}$. A simple extension then consists to consider also
\[
\overline{\Delta\pi_r} = \log\frac{1-\pi_r^{\rm test}}{1-\pi_r^{\rm train}}
\]
which will as well indicate privacy leaks in the underfitting regime when very negative.
In addition we expect the excess or deficit of sample below $u_r$ to have multiplicative effect on the 
probability $\pi_r^{\rm ref}$, so taking the $\log$ should scale linearly, hence qualitatively we  expect that both $\Delta\pi_r$ and $\overline{\Delta\pi_r}$ should scale linearly with the number of privacy leakage up to rank $r$, respectively in the over- and under-fitting regime.

\begin{table}[t]
\centering
\small
\setlength{\tabcolsep}{4pt}
\renewcommand{\arraystretch}{1.25}
\caption{\textbf{Summary of statements and assumptions.}}
\label{tab:assumptions}
\begin{tabularx}{\textwidth}{@{}c L{0.40\textwidth} L{0.40\textwidth} X@{}}
\toprule
\# & Statement &  Comments \\
\midrule

\multicolumn{3}{@{}l}{\emph{Nearest-neighbour distance distribution}}\\[2pt]

1 & $F(u,N)=1-\mathbb{E}_{x\sim\rho}\!\big[(1-F_x(u))^N\big]$
  & Exact if independence of the $N$ train samples.  \\

2 & $F(u,N)\leqslant 1-(1-F(u))^N$ & with equality iff $F_x$ independent of  $x$\\

3 & The rescaled minimum distance converges to a Weibull or Gumbel law
  & Domain-of-attraction condition on the lower tail of $F$ \\

\addlinespace[3pt]
\multicolumn{3}{@{}l}{\emph{Geometry of the data}}\\[2pt]

4 & Data lie on a $d$-dimensional manifold with $\rho$ smooth. 
  & Model assumption with small $x$ dependency\\

5 & Weibull with exponent $\alpha=d$. 
  & assuming \#4 and intrinsic dimension of the data $d=O(1)$. \\

6 & Gumbel with scale $\hat B$ independent of $x$
  & assuming \#4 and intrinsic dimension $d\gg1$.  \\

7 & The mixture over $x$ collapses onto a single EVD 
  & dominated by data from mostly dense regions \\

\addlinespace[3pt]
\multicolumn{3}{@{}l}{\emph{Order statistics and scoring }}\\[2pt]

8 & $P_{N,M}(u,r)$ has no error dependence with $M$
  & assuming independence of generated samples  \\

9 & corrections due to Train-train NN distances dependence negligible at large $N$  
  & vanishes as $O(1/\sqrt{N})$ and enters the estimation of $\hat G$ only.\\

10 & Error of the fit $\delta\hat G/\hat G=O(1/\sqrt{N})$.
  & Overall error of the fit including \#9 ; empirically assessed by the
    split replicates and binomial bands of 
    Figs.~\ref{fig:GoF_main} and~\ref{fig:GoF_splits} \\

\addlinespace[3pt]
\multicolumn{3}{@{}l}{\emph{Choice of distance }}\\[2pt]

11 & The fitted slope is asymptotically invariant across norm-derived distances
  & valide when $N\to\infty$ with $d$ finite \\

\bottomrule
\end{tabularx}
\end{table}

\section{\textcolor{black}{Dependencies in train-train distances vanishes as dataset size grows to infinity}}
\label{app:dependency_penrose}

There is a potential dependency between NN distances (only within the train data not between train-test or train-synth) due to the so-called triangle property of random geometric graphs ~\cite{Penrose2003-cy}. But these dependency vanishes when the number of samples $n$ within that set goes to infinity in a way that is fully characterized in ~\cite{Bickel1983-xd}.

For samples $\{X_1,\ldots X_n\} \in \R^d$ distributed with p.d function $f$ (typically uniformly bounded and continuous on his support)
Call $S(x,r)$ the sphere of radius $r$ centered at $x$ and defined with respect to an arbitrary distance. Call
$F(A) =: \int_A f(x) dx$ for any Borel set $A$ (e.g. $A= S(x,r)$ ), the probability to draw a sample in $A$.
Then call
$U_{jn} =: \exp(-n F(S(X_j,R_j))$   where $R_j$ is the NN distance to $X_j$ among the set of samples.
$W_{jn}  =:  \exp(-n f(X_j) V(R_j) )$  where $V(R_j)$  is the volume of the sphere of radius $R_j$.

The first statement is that both $U_{jn}$ and  $W_{jn}\sim U[0,1]$ at large $n$. In addition, calling $\hat H_n$ the empirical distribution
of the $W_{jn}$  and $Z_n(t) =: \frac{1}{\sqrt{n}}(\hat H_n(t) - E\hat H(t) ) $  for $0\le t\le 1$
is a zero mean Gaussian process whose variance is independent of $f$.

As a result, since the volume of the sphere is of the form
$V(r) = K_d r^d $ with $K_d$ a constant which depends on the choice of distance we can first conclude that $D_{jn} =: R_j n^{1/d}$ has a universal
distribution. Then for a function $h(x,\delta)$ uniformly bounded and almost everywhere continuous we have a central limit theorem for the sum
$\frac{1}{\sqrt{n}}\sum_{j=1}^n  h(x,D_{jn}) \to \mathcal N(\bar h,\sigma^2) $ in distribution with $\sigma$ finite. This basically means
that the fit function to the  empirical distribution of the $D_{jn}$ will have fluctuations of order $1/\sqrt{n}$ around the universal distribution.
\cf{More precisely these fluctuations concern the parameters of the fit which result as relative errors on the fit itself. In practice this means that the fit $\hat G(u)$ of the NN distribution  has fluctuations of order $\hat G(u)/\sqrt{N}$ which convert into fluctuations of  order 
$\mathcal O\Bigl(\frac{1}{\sqrt{N}}\Bigr)$ on the score~(\ref{eq:new_score}) which is a log ratio.
}

\section{About choosing distances}
\label{app:dist}

\subsection{Asymptotic equivalence of norm-derived distances}
\label{app:dist_equiv}

While the choice of a meaningful distance is \emph{a priori} crucial, we show nonetheless that in the aymptotic regime with infinitely many samples, the slope of the fit computed by our method based on Extreme Value Theory does not depend on the choice of the distance, provided it derives from a norm.

\newcommand{\X}{\mathcal{X}}

\noindent\textbf{Proof}

In finite dimension $p$ indeed, all norms are equivalent, that is, for any two distances $d$ and $d'$ derived from norms on $\mathcal{X} = \R^p$, 
$$\exists \alpha, \beta > 0, s.t. \forall \x_1, \x_2 \in \mathcal{X}, \;\;\;\;\; \alpha \,d(\x_1,\x_2) \leqslant d'(\x_1,\x_2) \leqslant \beta\, d(\x_1,\x_2)$$

This has consequences on nearest-neighbor distances: given a dataset $\D = \{ \x_i \}_{1\leqslant i \leqslant n}$ of points in $\X$, the distances of any given point $\x \in \X$ to the closest neighbor in the set $\D$ are similar. Indeed, noting $\delta(\x) = \min_{i\in\D} d(\x,\x_i)$ and $\delta'(\x) = \min_{i\in\D} d'(\x,\x_i)$, and $\x^*, \x^{*'}$ the associated closest points for $d$ and $d'$ respectively, we have:

$$\alpha\, \delta(\x) = \alpha\, d(\x, \x^*) \leqslant \alpha\, d(\x, \x^{*'}) \leqslant d'(\x, \x^{*'}) = \delta'(\x) \leqslant d'(\x, \x^*) \leqslant \beta\, d(\x, \x^{*}) = \beta\,\delta(\x)$$

hence $\;\alpha\, \delta(\x) \leqslant  \delta'(\x) \leqslant  \beta\,\delta(\x),$
and in particular:
$$\log\alpha + \log\delta(\x) \leqslant  \log \delta'(\x) \leqslant  \log\beta + \log\delta(\x)\;.$$

The histograms of nearest neighbor distances are thus quite similar, as histogram mass can be at most shifted by bounded values, namely $\log\alpha$ and $\log\beta$.

Now, if the range of distances (in log scale) on which the linear fit is performed (see Figure 1) is arbitrary large, the consequences of such bounded shifts on the fitted slope become arbitrarily negligible as well.

It turns out that when the number $M$ of samples increases, the nearest-neighbor distance histogram can be refined (since more points allow for more precise density estimation), and the region where the linear fit is performed gets extended towards smaller values (in log scale). More precisely, this region is proportional to  $[\frac{1}{M}, q]$ where $q$ is a fixed quantile and $\frac{1}{M}$ is the smallest quantum of probability available (corresponding to the weight of one sample in the dataset); in log scale this yields $[-\log M, \log q]$, a range that tends to infinity when the number of samples does as well. Q.E.D.

\subsection{Discussion about distances}
\label{app:dist_disc}

While all norms satisfy the property above, note that this is not the case of semi-norms. Indeed semi-norms yield 0 distance in certain directions, which implies $\alpha = 0$ and/or $\beta = +\infty$, breaking the equivalence and the asymptotic property. Note also that not all distances can be derived from norms.

The choice of a meaningful distance is crucial, as the privacy leaks are perceived under that lens. This said, as shown by the equivalence aymptotic property under certain conditions, EVT can detect leaks beyond the original intention: for instance in our application case with DNA and Hamming distance between SNPs, one could expect to detect only generated samples that are too close to one training sample (i.e. most SNPs are identical). In the meanwhile, an important concept in population genetics is the one of recombination (copying a part from an individual and then another part from another one). This yields a notion of similarity between individuals that does not seem related to the Hamming distance at first glance, as a recombination between 2 samples is far from each of the samples for that distance, in terms of number of SNPs that differ. However, a recombination is still visible in term of Hamming distance, as the distance to its original individuals is halved, w.r.t. typical distances (as half of the SNPs are identical). Statistically, in high dimensions, this is not probable, and will consequently be detected by the EVT approach.

\section{Assessing the quality of fits to the NN distributions}
\label{app:GoF_analysis}
\subsection{Goodness-of-Fit}

We perform a meta-analysis across datasets and modalities to empirically assess the extreme-value tail assumption. More specifically, we assess whether the empirical distribution of nearest-neighbor distances in each dataset is consistent with extreme value distributions. The procedure for FIG.\ref{fig:GoF_main}. Bottom. is as follows:

\textcolor{black}{Before computing nearest-neighbour distances, exact duplicate records are removed. The data are then partitioned into a training set and a held-out set. For the image and tabular datasets we use an $80/20$ split. For 1000 Genomes, we divide into equal parts. $N$ and $M$ denote the resulting numbers of training and held-out observations, respectively. We then fit
Weibull and Gumbel laws on the data-specific window $[q_1,q_2]$, keeping
the family with the smaller residual error. The selected initial windows $[q_1,q_2]$ are $[0.1\%,2\%]$ on CIFAR-10, $[0.1\%,1\%]$ on CelebA, $[0.1\%,10\%]$ on ImageNet, $[0.1\%,2.8\%]$ on UCI Adult, $[0.2\%,10\%]$ on UCI Bank and $[1\%,20\%]$ on 1000 Genomes, with $q_1$ held fixed throughout.}

\paragraph{\textcolor{black}{Binomial band.}}

The held-out test set provides a finite-sample reference for the
deviations visible in the P-P plot of Fig. Bottom. \ref{fig:GoF_main}. Consider any fixed distance $\delta$ and suppose that the fitted law assigns probability $p=\hat G(\delta).$ 
If this probability were correct, each independently drawn held-out observation
would have probability $p$ of having a test-to-train NN distance below
$\delta$. Consequently, the number $K_\delta = \#\{d^*_{TeTr}\leq\delta\}$ would satisfy $K_\delta\sim\operatorname{Binomial}(M,p).$

For each value of $p$ on the horizontal axis, we therefore compute the
$2.5$th and $97.5$th percentiles of $\operatorname{Binomial}(M,p)$ and divide
them by $M$:
\begin{equation}
    \left[
      \frac{Q_{0.025}\{\operatorname{Binomial}(M,p)\}}{M},
      \frac{Q_{0.975}\{\operatorname{Binomial}(M,p)\}}{M}
    \right].
    \label{eq:gof_binomial_band}
\end{equation}
This gives the orange $95\%$ binomial band in
Fig. Bottom. ~\ref{fig:GoF_main}, which thus represents finite-sample variability of the held-out observations for a fixed fitted probability.

\subsection{Consistency of the fit across random splits}
\label{app:gof_splits}

\textcolor{black}{
The previous diagnostics consider one particular training \& held-out partition. We additionally ask whether the fitted tail law depends on any particular partition.}

\textcolor{black}{For this purpose, we repeat the estimation on $R=10$ independent random splits. For each replicate $r$, the de-duplicated dataset is shuffled with its own seed and divided $80/20$ into a fitting training set and a held-out test set. We compute the nearest-neighbour distances within the fitting set, fit $\hat G_r$ on that dataset's window $[q_1,q_2]$ with the family with lowest residual error, and then evaluate the frozen $\hat G_r$ on the test-to-train 1NN distances, which played no part in the estimation. Since the dataset is fixed, shuffling with different seeds creates random splits, however these will share samples between seeds.}

\begin{figure*}[!h]
    \centering
    \includegraphics[width=\textwidth]{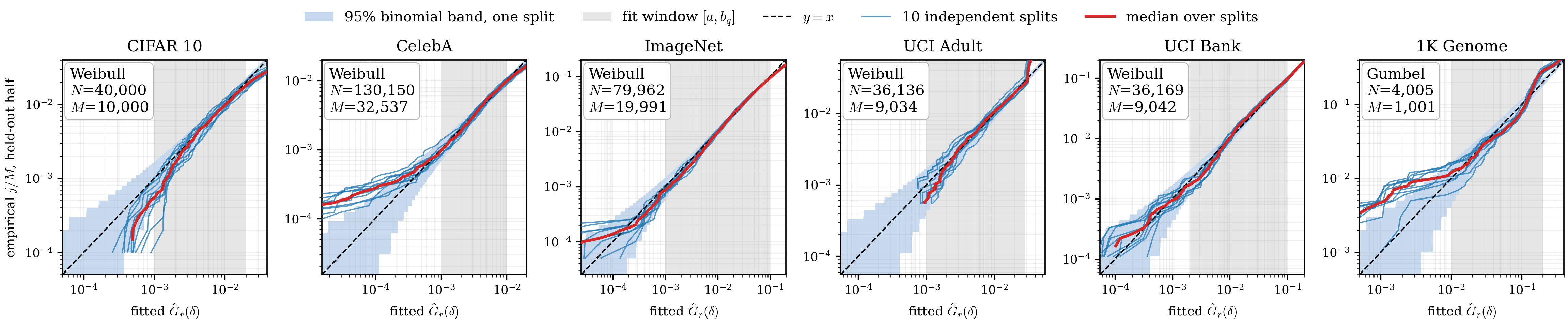}
    \caption{\textcolor{black}{\textbf{Reproducibility of the EVT fit across random splits.}
    Blue curves compare the resulting fitted probabilities
    $\{\hat G_{r}(\delta)\}_\delta$ with the empirical cumulative probability in the
    corresponding held-out subset, the red curve denotes the median over
    partitions. The grey region denotes the dataset-specific fitting interval
    and the shaded band is the $95\%$ binomial band for the
    held-out sample size of one replicate. Similar curves across partitions indicate
    that the fitted tail law is reproducible with respect to the random data
    split.}}
    \label{fig:GoF_splits}
    \vspace{-5mm}
\end{figure*}

\subsection{\textcolor{black}{Distribution of $\Delta p_r$ under a no-leakage null}}

\begin{figure*}[!h]
    \centering
    \includegraphics[width=0.9\textwidth]{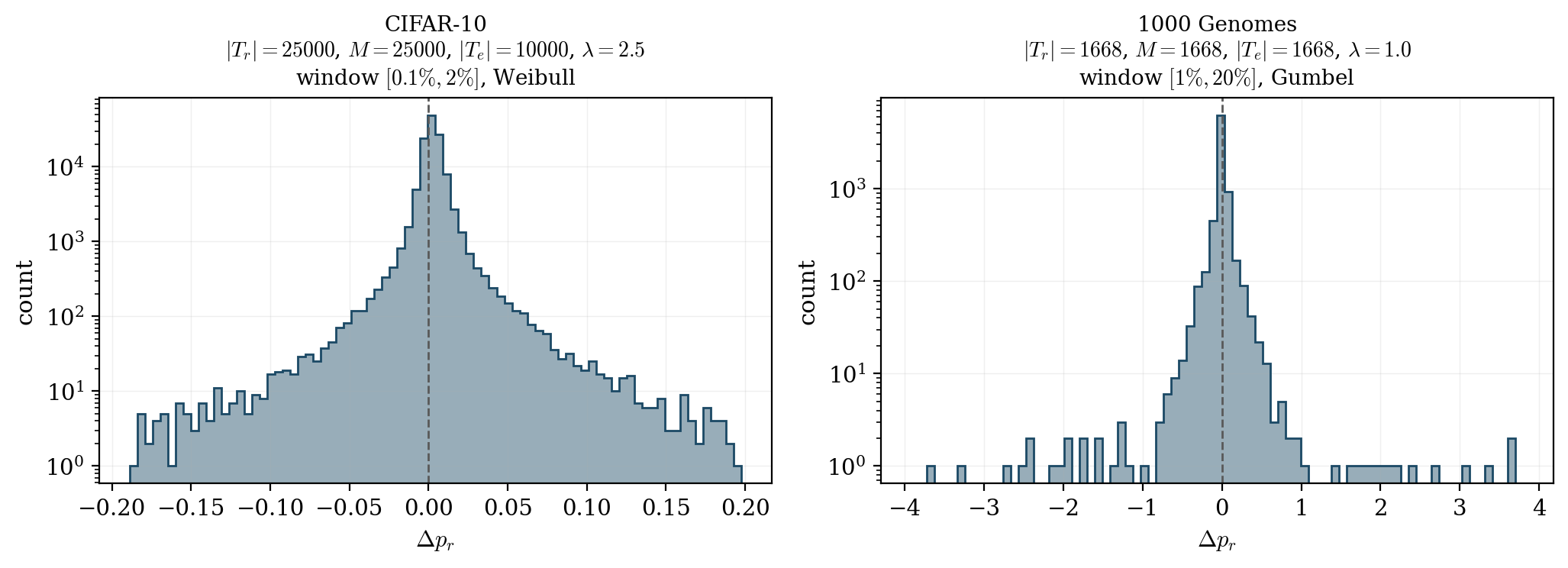}
    \caption{\textcolor{black}{Distribution of the per-sample score $\Delta p_r$ under a no-leakage null, pooled over $R=5$ random partitions. For each replicate, we split the dataset into three disjoint, exchangeable real subsets playing the roles of training, synthetic, and held-out test set. We then compute the scores and repeat this process $R$ times. \textbf{Left.} DINOv2 embeddings of CIFAR-10 are used to computed distances, with a fit window of $[0.1\%,2\%]$. \textbf{Right.} 1000 Genomes, Hamming distance, fit window $[0.1\%,2\%]$. Dashed line represents$\Delta p_r=0$.}}
    \label{fig:noleakage_null_score_distrib}
\end{figure*}

\textcolor{black}{To visualize the null distribution underlying our calibration results, Figure.~\ref{fig:noleakage_null_score_distrib} shows the distribution of per-sample score $\Delta p_r$ under a no-leakage split on CIFAR-10 and 1000 Genomes. In both cases the distribution is unimodal, symmetric, and centred around zero, with no anomalous excess mass in the negative tail, i.e. the region a real leak would populate.} 

\textcolor{black}{This indicates that PRIVET extends naturally to mixed data types (categorical + continuous): in our experiments, euclidean distance on preprocessed features (normalization and one-hot encoding) was sufficient, and alternative distances for heterogeneous data types exist in the literature \cite{Meeus_2024}.}

\clearpage
\section{Controlled experiments on synthetic data}
\subsection{Genetic data}
\label{1KG_data}

The 1000 Genomes (1KG) Project is a large-scale public resource of human genome sequences designed to represent the global genetic diversity of the human population by sampling individuals from distinct populations across multiple continents \cite{the1000genomesprojectconsortiumGlobalReferenceHuman2015}. The dataset used in this study comprises 2504 diploid individuals corresponding to 5008 phased haplotypes (i.e. for each individual there is one reconstructed haplotype coming from the mother and one from the father), each of  65,535 contiguous SNPs spanning chr1:534247-81813279 (approximately 80 megabase pairs) within the Omni 2.5 genotyping array framework. The data is organized as follows: rows represent phased haplotypes (one of the two chromosome copies inherited from a parent, with known ordering of variants), while columns indicate allele positions, encoded as 0 when the nucleotide matches the reference genome (GRCh37), and 1 when it differs (i.e., a point mutation). This results in a binary matrix of aligned sequences.

With the ground-truth datasets introduced in Main. Subsection.~\ref{genetic_data}, we conducted a comprehensive evaluation using a diverse set of metrics, that quantify overfitting in generative models (section \ref{sec:related_work}).
For $\mathcal{AA_{TS}}$ \cite{YALE_2020}, FLD \cite{NEURIPS2023_68b13860} and PQMass, we considered the generalization gap \cite{lemos2025pqmass}, i.e., $Score(\mathcal{D}_{test},\mathcal{D}_{synth}) - Score(\mathcal{D}_{train},\mathcal{D}_{synth})$ where $Score()$ is one of these metrics. $Privacy \text{ }Loss$ (i.e. the generalization gap of $\mathcal{AA_{TS}}$ along with that of  PQMass is strictly positive when overfitting is detected, while those of FLD and $C_T$ \cite{meehanNonParametricTestDetect2020} are strictly negative).

\begin{figure*}[!h]
\centering
\includegraphics[width=1\textwidth]{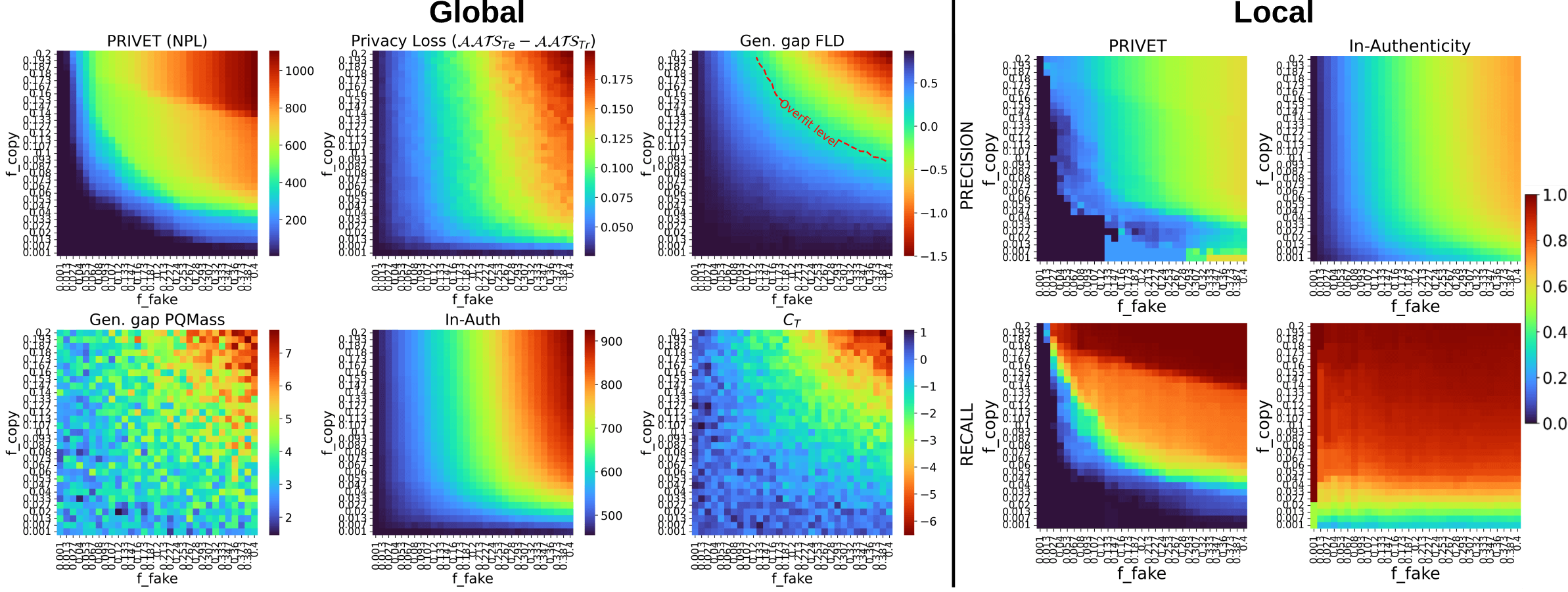}
\vspace{-0.6cm}
\caption{\textbf{Global and local privacy scores for 65,535 SNPs and 1,668 samples.} 
\textbf{Global:} From top left to bottom right—number of privacy leaks (NPL \textcolor{black}{Eqs.~\ref{eq:global-scores-inline}}) from PRIVET, generalization gaps of $\mathcal{AATS}$, FLD, and PQMass, In-Authenticity, and $C_T$, plotted over $f_{\rm fake}$ vs. $f_{\rm copy}$. Color indicates metric values; for FLD, the overfit level defines the contour above which overfitting is detected. \textbf{Local:} Precision (top) and recall (bottom) are reported for local privacy metrics. \textcolor{black}{PRIVET score $\Delta \pi_r$ (Eqs.~\ref{eq:priv_index}) was thresholded with $\tau = -3$.}}
\label{fig:privacy_map_65K}
\end{figure*}

Fig.~\ref{fig:privacy_map_65K} compares the behavior of the different global scores when both $f_{\rm fake}$ and $f_{\rm copy}$ are varied. We observe that $C_T$, the generalization gap of FLD, and PQMass exhibit limited sensitivity to these changes, notably $C_T$ and the generalization gap appear noisy. In the case of PQMass, the observed noise may also stem from its reliance on a single set of reference points used to create the tessellations. A potential solution to mitigate this noise would be performing multiple repetitions varying the reference points (e.g., 20 as performed by the original authors, see Fig. 8 in ~\cite{lemos2024pqmass}), but it would be computationally heavy for each $(f_{\rm fake}, f_{\rm copy})$ configuration in our case. For gen. gap FLD, the zone above the overfit contour line where a leak signal would be triggered corresponds to already large values of $f_{\rm fake}$.
By contrast, the number of privacy leaks (NPL) detected by PRIVET looks less noisy and more sensitive both to $f_{\rm copy}$ and $f_{\rm fake}$.
In particular, PRIVET (NPL, \textcolor{black}{Eqs.~\ref{eq:global-scores-inline}}) exhibits a sharp threshold separating a phase where little or no privacy leak is detected (dark domain) at low values of both $f_{\rm fake}$ and $f_{\rm copy}$
from the privacy leak phase (colored domain), with as a bonus a precise indication of the number of synthetic data that leaks information. Privacy loss and the negated Authenticity score (In-Auth) also appear highly sensitive to privacy leakage. However, Privacy loss provides only a global (i.e., dataset-level) assessment and is therefore not interpretable, as it does not identify which specific samples are concerned with information leakage. 
Although In-Auth demonstrates strong precision and recall across both privacy leak axes, PRIVET achieves higher recall for $f_{\rm copy} \geq 0.15$, as illustrated in Fig.~\ref{fig:privacy_map_65K}. While In-Auth appears overall stronger, it does not account for dataset size and is therefore highly sensitive to statistical fluctuations. As a result, a synthetic sample may appear spuriously close to a training point purely by chance (see FIG.~\ref{fig:privacy_map_glocal_low_sample_size}). Moreover, by construction, In-Auth cannot detect privacy leakage in the underfitting regime.

\subsection{Low sample size}
\label{app:controlled_leak_low_sample}
In addition to the experiments described in the main text, we illustrate the robustness of PRIVET in a low-sample-size setting.
These data are real genetic data corresponding to binary sequences of size $33602$ retrieved from \citet{the1000genomesprojectconsortiumGlobalReferenceHuman2015}. Distance between
samples is simply the Hamming distance. We partition these $198$ samples into three random subsets each of equal size $66$  and arbitrarily assign these to train, test and synthetic datasets, the latter corresponding to the output from a fictional generative model. The x and y-axis corresponding respectively to $f_{\rm fake}$ and $f_{\rm copy}$ are used to construct pseudo-synthetic data as presented in Main. Section.~\ref{genetic_data}.

\begin{figure*}[!h]
\vspace{-0.2cm}
\centering
\includegraphics[width=1\textwidth]{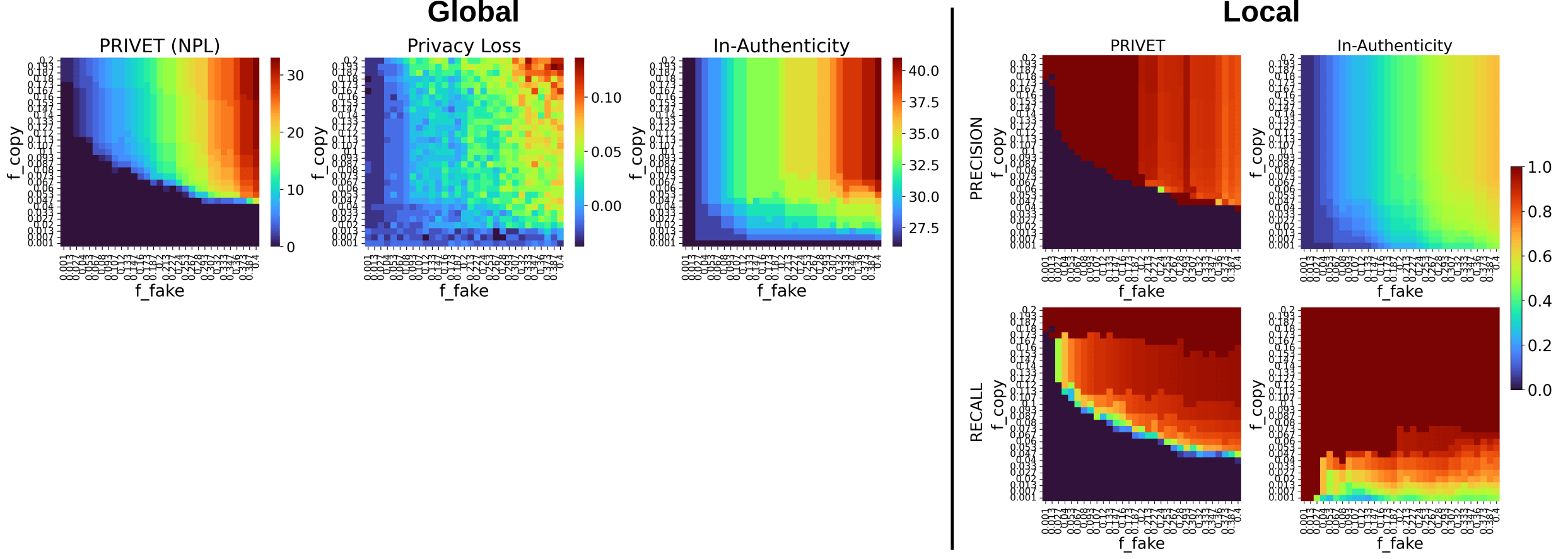}
\vspace{-0.2cm}
\caption{\textbf{Global and local privacy scores for 33,602 SNPs and 66 samples.} \textbf{Global:} From left to right—number of privacy leaks (NPL) from PRIVET, generalization gaps of $\mathcal{AATS}$ (i.e. Privacy loss) and In-Authenticity score, plotted over $f_{\rm fake}$ vs. $f_{\rm copy}$. \textbf{Local:} Precision (top) and recall (bottom) are reported for local privacy metrics.}
\label{fig:privacy_map_glocal_low_sample_size}
\vspace{-0.4cm}
\end{figure*}

Fig.~\ref{fig:privacy_map_glocal_low_sample_size} compares the behavior of the different global scores when both $f_{\rm fake}$ and $f_{\rm copy}$ are varied. Global results are not shown for FLD ~\cite{NEURIPS2023_68b13860}, PQMass ~\cite{lemos2024pqmass} and $C_T$ ~\cite{meehanNonParametricTestDetect2020} due to execution failures caused by small sample size that cannot be handled by these metrics. We observe limited sensitivity and noisy behavior for Privacy loss score of $\mathcal{AATS}$ in contrast to the scenario where sample size is larger (c.f. Main. Fig.~\ref{fig:privacy_map_65K}). PRIVET recovering the number of privacy leaks (NPL) exhibits strong sensitivity to both axes. A distinct boundary separates regions of no detection from detection, beginning at $18\%$ copied SNPs (from the closest neighbors) and with as few as $0.1\%$ leaked samples (i.e., 1 synthetic samples), and extending to $40\%$ leaked synthetic samples with as little as $4.7\%$ copied SNPs. In-Auth is highly responsive to privacy leakage signals and achieves strong recall, however, this comes at the cost of many false positives, resulting in substantially lower precision than PRIVET in the intermediate and high leakage regimes.

\begin{figure*}[t]
\centering
\includegraphics[width=1\textwidth]{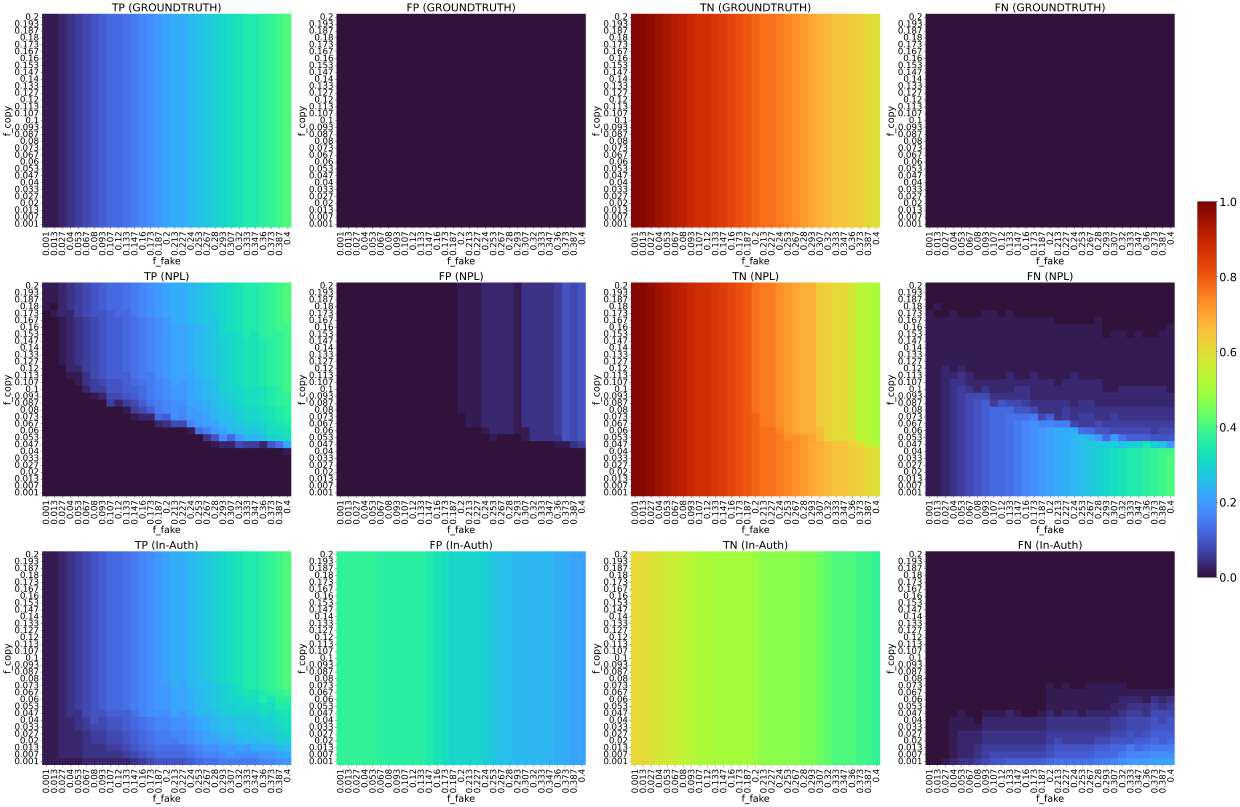}
\caption{\textbf{Classification results for 33,602 SNPs and 66 samples.} The columns represent: True Positives (TP), False Positives (FP), True Negatives (TN), and False Negatives (FN). The rows represent the groundtruth, PRIVET (NPL) and In-Authenticity score. Privacy leaks are reported as a percentage of the synthetic data. $(f_{\rm fake},f_c)$ range from $(0.1\%,0.1\%)$ to $(40\%,20\%)$.}
\label{fig:privacy_map_groundtruth_low_sample_size}
\end{figure*}

Dissecting further, we report true/false positives and negatives for the groundtruth, PRIVET (NPL) and In-Authenticity, as these are the only methods providing binary (leak/no-leak) predictions at the sample level (Fig. ~\ref{fig:privacy_map_groundtruth_low_sample_size}). The first row shows the groundtruth, defined by the values of $f_{\rm fake}$. In the second row, PRIVET (NPL) exhibits few false positives on second column. In contrast, the false positive map produced by In-Authenticity remains flat along $f_{\rm copy}$ axis with rates substantially higher than for PRIVET (NPL).

\subsection{Classification results for 65,535 SNPs and 1,668 samples}

\begin{figure*}[!h]
\centering
\includegraphics[width=1\textwidth]{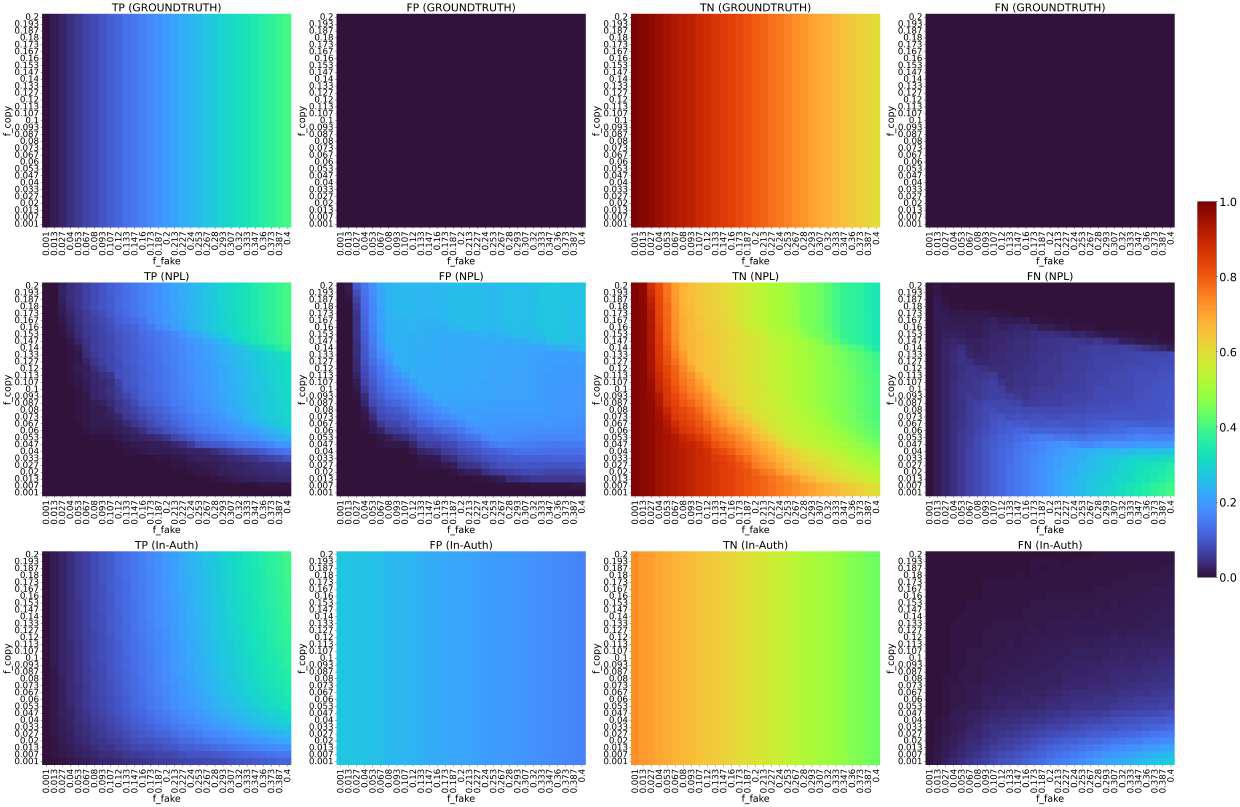}
\caption{\textbf{Privacy maps for 65,535 SNPs and 1668 samples.} The columns represent: True Positives (TP), False Positives (FP), True Negatives (TN), and False Negatives (FN). The rows represent the groundtruth, PRIVET (NPL) and In-Authenticity score. Privacy leaks are reported as a percentage of the synthetic data. $(f_{\rm fake},f_c)$ range from $(0.1\%,0.1\%)$ to $(40\%,20\%)$.}
\label{fig:privacy_map_groundtruth_65K}
\end{figure*}

In Fig.~\ref{fig:privacy_map_groundtruth_65K} we display PRIVET (NPL) and In-Auth true positives and false positives maps for the larger sample size genetic dataset (65,535 SNPs and 1,668 samples) Note that the sum of the two maps (TP+FP) yields the PRIVET (NPL) and In-Auth scores presented in Main.Fig.~\ref{fig:privacy_map_65K}. Once again, we can see the tendency of In-Auth of overestimating privacy leakage.

\subsection{Settings for benchmarked metrics} \label{sec:metric-settings}
For all controlled experiments and all metrics, including $C_T$, FLD, Authenticity Score, and the Privacy Loss of $\mathcal{AATS}$, we applied default parameter values. We relied on the official GitHub implementation from~\cite{NEURIPS2023_68b13860} for all metrics, except PQMass and Privacy Loss, which were implemented separately.

For PQMass, we set \verb|re_tesselation = 1000|, following author's recommendation that higher values yield improved performance.

A normalization preprocessing step is applied internally by FLD. However, when applied to genetic data, this step fails in the presence of fixed sites—SNP positions where all individuals share the same allele (0 or 1)—leading to errors. To address this, we excluded fixed sites from the test data, and then removed the corresponding positions from the train and synthetic sets as well. This resulted in the removal of approximately 200 SNPs per dataset.

For PRIVET, we fitted the NN distances between train-train samples with a Weibull distribution for all datasets. When no mode is observed on the eCDFs, we typically fit on the partition of the data corresponding to the first $1\%$ to $20\%$ of the sorted NN distances. When multi-modal structure is observed, the partition typically ranges from $0.1\%$ to 2 or $5\%$, depending on the specificity of the first mode on which we want to apply the fit.

\clearpage
\subsection{Score refinement}

\paragraph{Decimation}
 Although to a substantially lesser extent than In-Auth, PRIVET (NPL) still produces false positives. To address this issue, synthetic samples with privacy scores below the threshold should be sequentially removed, with the scores of the remaining samples recomputed at each step to ensure a proper cleanup of the synthetic data. This additional decimation step adds a cost of $\mathcal{O}(kN)$ where $N$ is the number of synthetic samples and $k$ is the number of detected privacy leaks among them. We compared PRIVET with and without decimation (Fig.~\ref{fig:local_decimation}). The excess of small synthetic to train NN distances due to leakage artificially inflates the remaining eCDF, i.e. it artificially increases the rank of the remaining synthetic samples. This might in turn impact $\Delta \pi_r $ score. By progessively removing the synthetic privacy breaches, we observe a notable improvement in precision, albeit with a decrease in recall. Examining the F1 score, the decimated version of PRIVET shows reduced sensitivity to the copy fraction $f_{\rm copy}$, while achieving higher scores when both $f_{\rm fake}$ and $f_{\rm copy}$ are large. Given the context of privacy leakage detection, we argue that false negatives are more critical than false positives.

\paragraph{Implicit decimation}

\begin{figure*}[!h]
\centering
\includegraphics[width=1\textwidth]{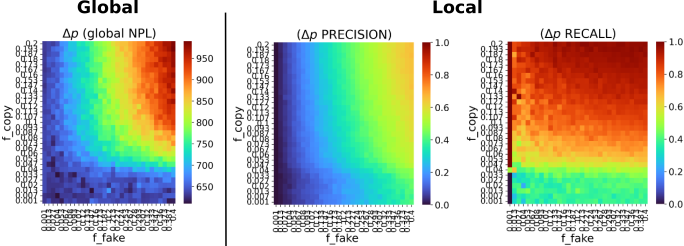}
\caption{\textbf{PRIVET local scores with implicit decimation (Eq.~\ref{eq:new_score}).} Controlled experiment on genetic data: 65,535 SNPs and 1,668 samples.}
\label{fig:local_implicit_decimation}
\end{figure*}

\begin{figure*}[!h]
\centering
\includegraphics[width=1\textwidth]{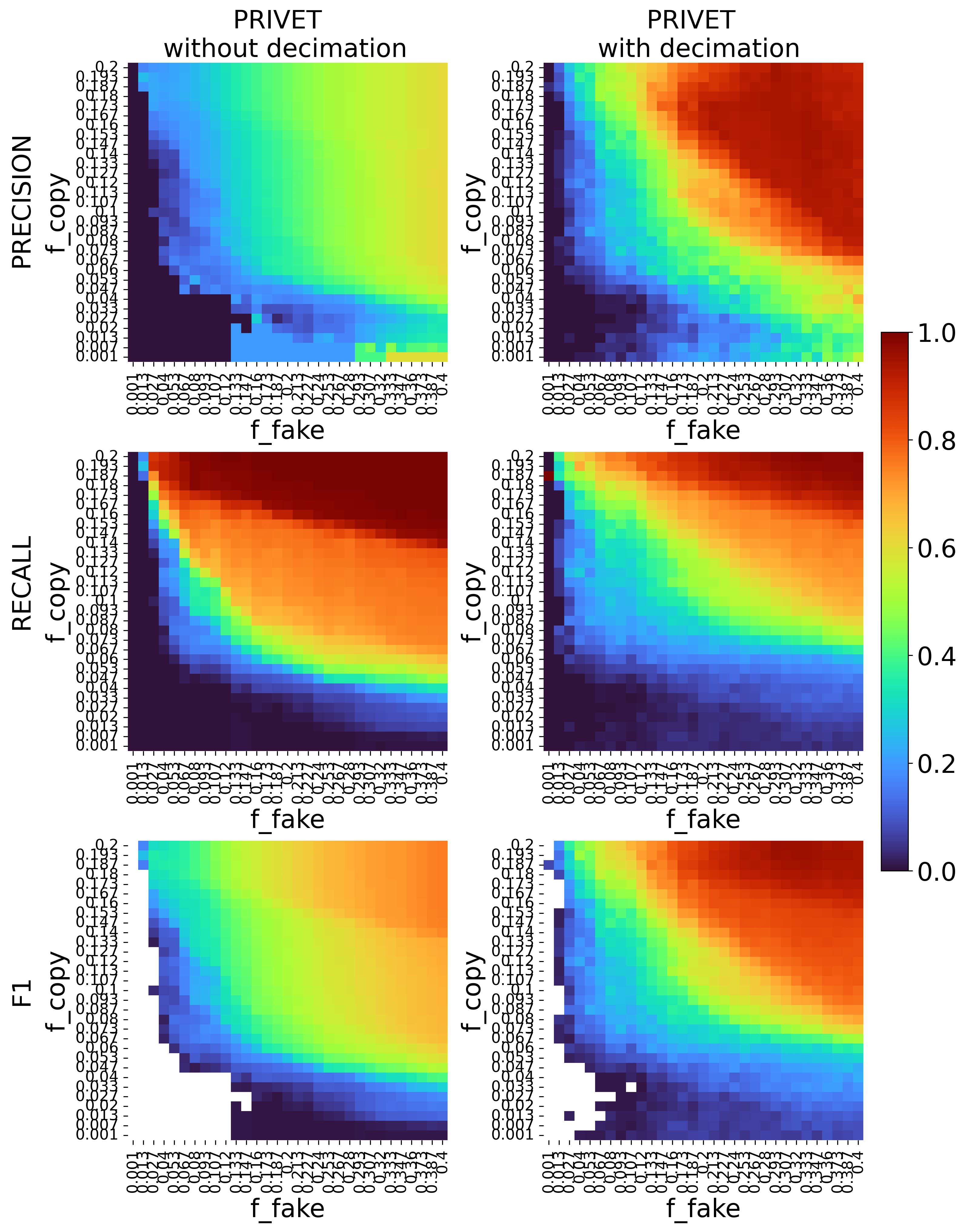}
\caption{\textbf{PRIVET local scores without and with decimation.} Controlled experiment on genetic data: 65,535 SNPs and 1,668 samples. Rows corresponds to precision, recall and F1 score respectively, plotted over $f_{\rm fake}$ vs. $f_{\rm copy}$. Columns correspond to PRIVET without decimation (see local privacy maps in Main.Fig.~\ref{fig:privacy_map_65K} and PRIVET with decimation. White pixels in F1 score correspond to nan value where precision and recall are both equal to 0.}
\label{fig:local_decimation}
\end{figure*}

In \textbf{FIG.\ref{fig:privacy_map_65K}.Local}, the recall of PRIVET appears to vary with the fraction of leaked samples, which is an undesirable dependency. By contrast, the individual privacy leak score defined in Eq.~\ref{eq:new_score} (Appendix~B) corrects for this effect while simultaneously improving upon the raw PRIVET score in \textbf{FIG.\ref{fig:privacy_map_65K}.Local}. With this alternative formulation, PRIVET achieves performance comparable to In-Auth (\textbf{FIG.~\ref{fig:local_implicit_decimation}}), creating smoother detections in the low $(f_{\rm fake},f_{\rm copy})$ region. Nonetheless, turning this continuous score into binary "leak" or  "not leak" flag requires careful threshold fine-tuning. In our experiments, we used a threshold of $-0.00115$ corresponding to a pareto front on a Precision-Recall plot.

\clearpage

 \section{Image data}

For the copycat experiment (c.f. Main.Section.~\ref{main_sec_image_data}), we used PyTorch's CIFAR10 train and test. We used synthetic data generated by SOTA diffusion model on CIFAR10 ~\cite{pfgmpp} that was provided by ~\citet{stein} and down-sampled it to 10,000 images instead of the 100,000 available. The down-sampling was achieved by taking the first 1,000 images of each class. \textcolor{black}{The pseudo-synthetic data is created by drawing $\lfloor \beta N \rfloor$ training examples uniformly at random and combining them with $N - \lfloor \beta N \rfloor$ synthetic examples (where $N$ is the size of the synthetic set). The experiment is repeated over five seeds, each redrawing the reference set $Tr$, the holdout $Te$ and the injected samples. The reported curves are means over seeds and bands give the min-max envelope. Variability due to the generator itself is not accessible, as the synthetic samples are a single released set from \cite{stein}}


For the computer vision transform, we used the following:
\begin{itemize}
    \item \texttt{Posterize(bits=5)}: Applies bit-level quantization by reducing each color channel to 5 bits using \texttt{torchvision.transforms.functional.posterize}, simulating color degradation.
    \item \texttt{CenterCrop(28) + Pad(2)}: Crops the center 28$\times$28 region of the input image and pads it by 2 pixels on each side, preserving input shape and emphasizing central image content.
    \item \texttt{JPEGQuality(quality=75)}: Re-encodes a \texttt{PIL.Image} using JPEG compression at quality level 75 via an in-memory \texttt{BytesIO} buffer, simulating lossy compression; used within a \texttt{torchvision.transforms.Compose} pipeline.
    \item \texttt{ElasticTransform()}: Applies elastic deformation to the image using \texttt{torchvision.transforms.ElasticTransform} with default parameters, introducing spatial distortions that simulate realistic image warping.
\end{itemize}

\texttt{fld.features.DINOv2FeatureExtractor} from FLD github was used to extract DINOv2 embeddings. Default value corresponds to ViT-B/14 distilled without registers. In practice we observed similar behavior for embeddings produced by ViT-B/14 and ViT-L/14 models, we thus used the former for computational efficiency reason.

For PRIVET, we fit the first $0.1\%$ to $2\%$ of the NN euclidean distances between train-train DINOv2 embeddings with a Weibull distribution. This corresponds to the first mode observed on the CDF. \textcolor{black}{The fitted law $\hat G_{d^*_{TrTr}}$ is evaluated in two regimes. Below the fit
window we extrapolate the fit. Above the fit window we read $\hat G$ off the empirical CDF of $d^*_{TrTr}$ instead of extrapolating. Indeed, data leakage is considered as an extreme event that happens in the lower tail, however the copycat's experiment design goes up to 100\% of leakage which is not the fit intended use. As train and test set have different sizes, a rescaling of the distances in the distribution of NN from synthetic-to-test samples has to be applied to correct for different densities. Instead of applying a theoretically derived renormalization factor, we consider an experimental approach that accounts for the regularities specific to the dataset. We define a monotone map estimated from data, $Q = \hat G_{d^*_{TrTr}}^{-1} \circ \hat G_{d^*_{TeTe}}$, and set $\delta^{STe} \leftarrow Q(\delta^{STe})$. The scoring is unchanged and still uses the single train fit. }


For PQMass, we set \verb|re_tesselation = 1000|, following recommendations that higher values yield improved performance.

$\beta$ fraction of copied training samples are in
$\{0, 0.01, 0.05, 0.1, 0.2, 0.3, 0.5, 0.8, 1\}$.

For wavelet embeddings we used \verb|wavelet = "sym5"| with \verb|max_level = 3|, resulting in feature vectors of dimension 23,232 for CIFAR-10 images with shape $(3, 32, 32)$ (channels, height, width).  


\textbf{Settings for benchmarked metrics} The settings were set similarly in the image and the genetic controlled experiments. Details are described in section \ref{sec:metric-settings}.

\begin{figure*}[!h]
\centering
\includegraphics[width=1\textwidth]{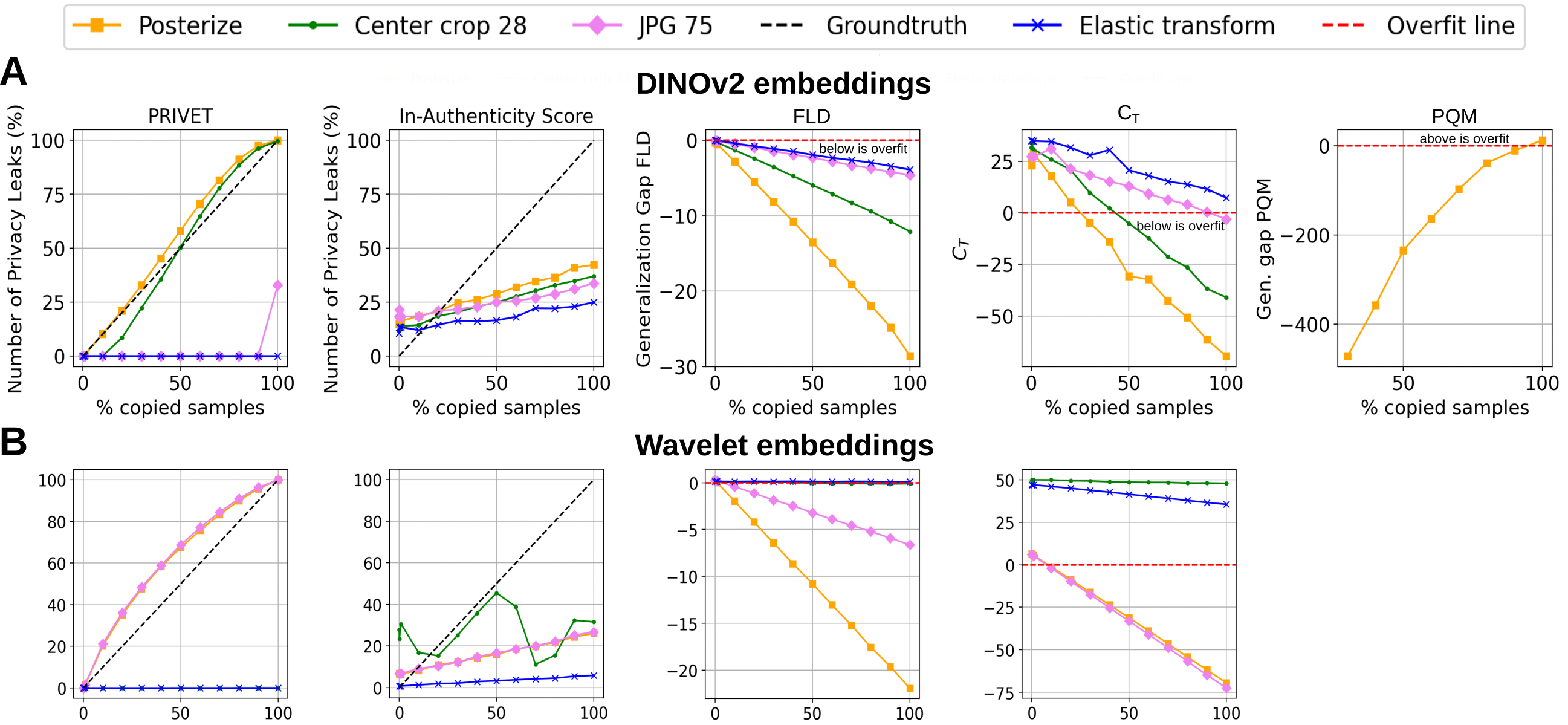}
\vspace{-0.5cm}
\caption{\textbf{Copycat experiment.} Several computer vision transformations (posterize, center crop 28, JPG 75, Elastic transform) are applied on CIFAR10 synthetic and training samples to generate a pseudo-synthetic dataset. x-axis represents the fraction of synthetic samples replaced with training ones; y-axis reports the value of the corresponding metric. For PRIVET and In-Authenticity, we report the percentage of identified privacy leaks. In the case of PRIVET, we use $\mathrm{NPL}$ from Eqs.~\ref{eq:global-scores-inline}. For Gen. gap (FLD) and $C_T$, overfitting is detected when the metric falls below the horizontal dashed line, whereas for Gen. gap (PQMass), overfitting is indicated when the metric exceeds the threshold. The top row presents results using DINOv2 (ViT-B/14 distilled without registers) embeddings of CIFAR10, while the bottom row shows results using wavelet packet coefficient space (symlet5, level=3). PQMass results are omitted for several transformations due to divergence, and are not shown for wavelet embeddings due to the high dimensionality, which caused excessive CPU time or memory limitations on GPU.}
\label{fig:copycat_dinov2_app}
\vspace{-0.5cm}
\end{figure*}

\begin{figure*}[!h]
\centering
\includegraphics[width=0.9\textwidth]{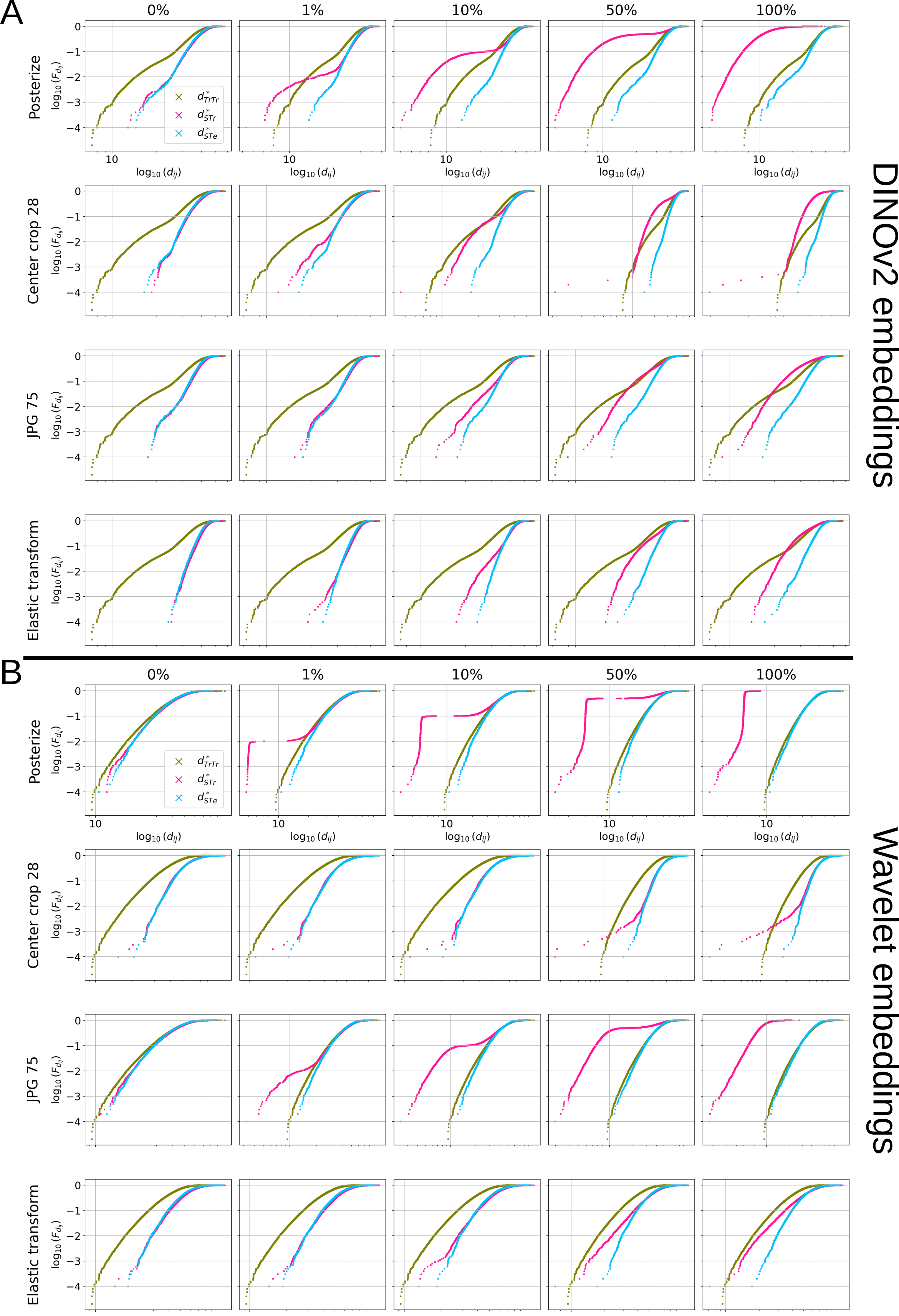}
\caption{\textbf{eCDFs for copycat experiment.} Rows are posterize, center crop 28, JPG75 and elastic transform respectively. Columns are fraction of copied samples from train into synthetic set (transformed beforehand) corresponding to $0, 1, 10, 50, 100\%$ respectively. \textcolor{olive}{$d_{TrTr}^*$}, \textcolor{magenta}{$d_{STr}^*$}, and \textcolor{cyan}{$d_{STe}^*$} denote the distributions of nearest-neighbor distances from train to train, synthetic to train, and synthetic to test, respectively. The x- and y-axes represent the distances and the eCDF, respectively, and are both shown on a $\log_{10}$ scale. \textbf{A.} eCDFs using DINOv2 embeddings. \textbf{B.} eCDFs using wavelet features.}
\label{app:cdf_copycat}
\end{figure*}

The underestimation of PRIVET in detecting privacy leaks in certain cases (see Main Fig.\ref{fig:copycat_dinov2}) can be better understood by examining the eCDFs shown in Fig.\ref{app:cdf_copycat}. In these instances, the synthetic-to-train eCDFs appear underfit, even though separation of synthetic-to-train and synthetic-to-test eCDFs indicates privacy leakage that should be detected by $\overline{\Delta\pi_r}$ and the other scores described in Appendix \ref{sec:various-scores}.

\clearpage
\clearpage

\subsection{\textcolor{black}{Low data regime}}

\begin{figure*}[!h]
\centering
\includegraphics[width=1\textwidth]{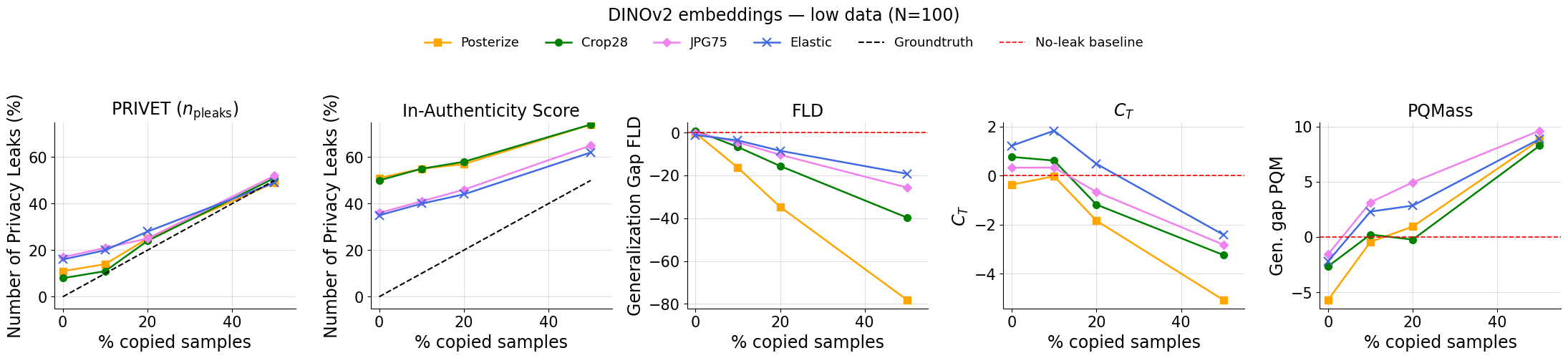}
\caption{\textcolor{black}{\textbf{Copycat experiment in low-data regime.} PRIVET's $n_{\mathrm{pleaks}}(r)$ score is reported (equation \ref{eq:new_pleaks}). CIFAR10 is subsampled to 100 images (10 samples per class)}}
\label{fig:copycat_dinov2_lowdata}
\end{figure*}

\begin{figure*}[!h]
\centering
\includegraphics[width=1\textwidth]{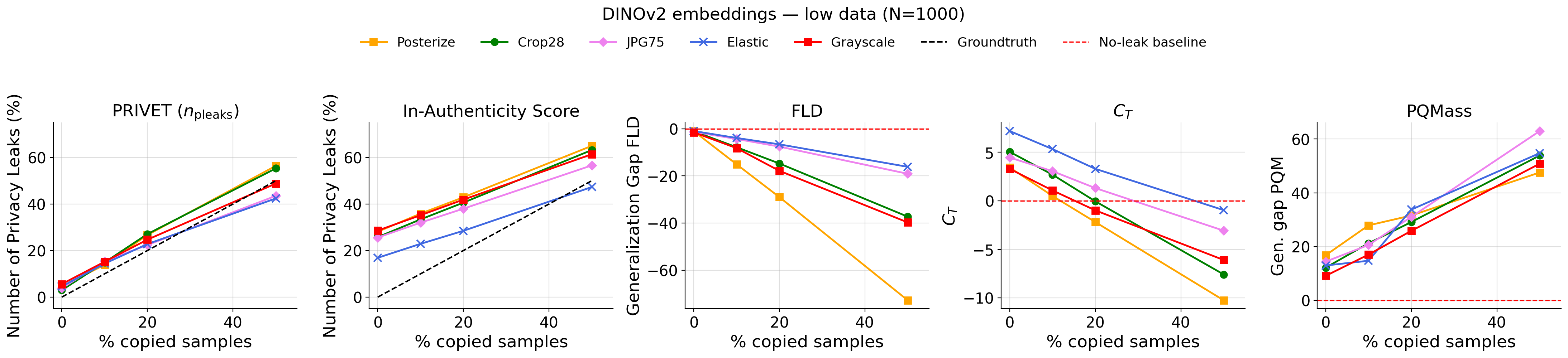}
\caption{\textcolor{black}{\textbf{Copycat experiment in low-data regime with grayscale transformation.} PRIVET's $n_{\mathrm{pleaks}}(r)$ score is reported (equation \ref{eq:new_pleaks}). CIFAR10 is subsampled to 1000 images (100 samples per class)}}
\label{fig:copycat_dinov2_lowdata_grayscale}
\end{figure*}

\textcolor{black}{We perform a low-sample experiment of the copycat experiment on CIFAR10 with samples across train, test and synthetic set to evaluate the robustness of PRIVET in that regime on image data (FIG.~\ref{fig:copycat_dinov2_lowdata}). Additionally, we also show that a color-to-grayscale copy of a training sample is within PRIVET's detection scope (FIG.~\ref{fig:copycat_dinov2_lowdata_grayscale}).}

\section{\textcolor{black}{Membership inference: what proximity signals reveal, and their limits}}  \label{sec:membership}

\begin{figure*}[!h]
\centering
\includegraphics[width=1\textwidth]{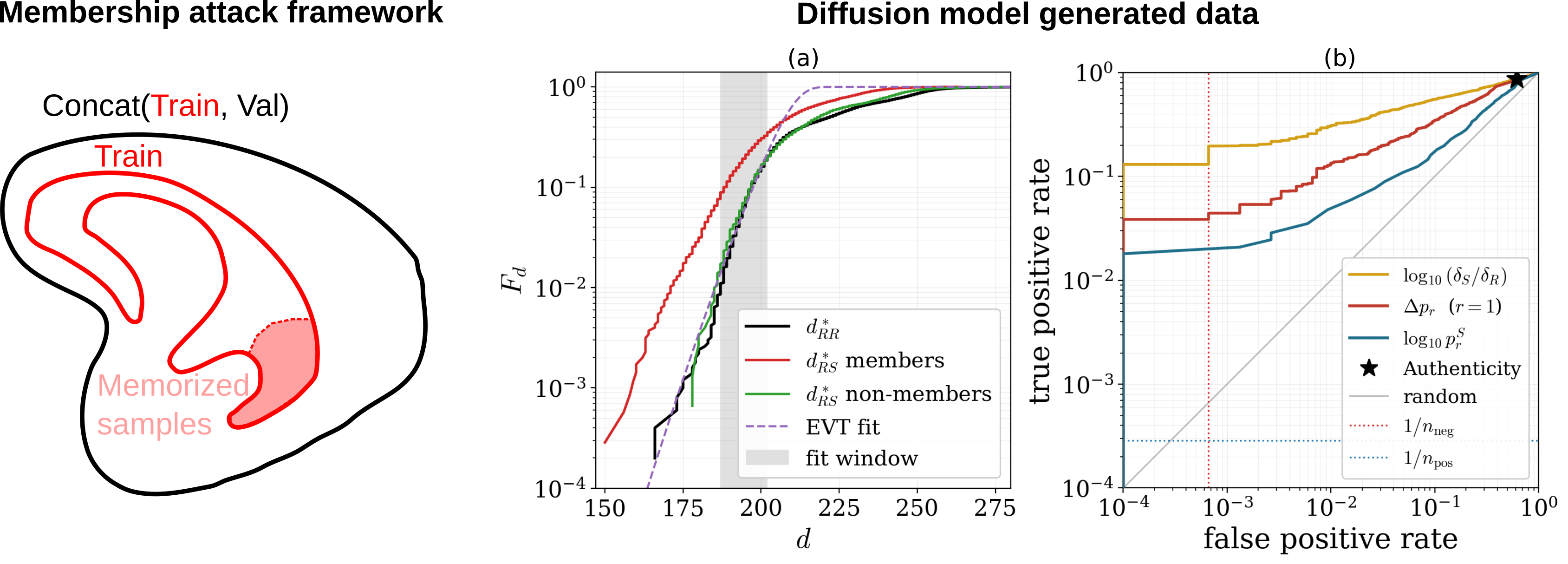}
\caption{\textcolor{black}{\textbf{PRIVET can be turned into a membership attack.} The reference data is created by merging train and validation set. The goal is to identify the training samples among the overall database. 
\textbf{(a).} eCDF for 1NN distances of reference pool to reference pool (black), training samples of the reference pool to synthetics (red), validation samples of the reference pool to synthetics (green).  \textbf{(b).} Membership inference ROC curves for diffusion model generated data and the different scores: log-ratio of distance (yellow), $\Delta p_{r=1}$ (red) and $\log p^S_r$ (blue). Floors corresponding to number of negatives (vertical red dashed line) and number of positives (horizontal blue dashed line) are shown. Grey diagonal corresponds to random classification. The retained diffusion model was trained for $12{,}500$ epochs and reveal an overfitting regime. The training and validation sets contain $3{,}500$ and $1{,}500$ samples respectively, and the DDPM synthetic set contains $10{,}000$ samples.}}
\label{fig:membership_attack}
\end{figure*}

Here we consider a scenario where we have a large public dataset and synthetic data generated by 
a private model, for which we want to identify in the public set which examples have been used to 
train the model. We performed the attack on a real-case scenario consisting of a diffusion model trained to generate genetic data. More specifically, the discrete diffusion model was trained on 3500 training samples of 805 highly differentiated SNPs ~\cite{colonnaHumanGenomicRegions2014} coming from ~\citep{the1000genomesprojectconsortiumGlobalReferenceHuman2015}. We then generated samples using a checkpointed  model at epoch $=12{,}500$ corresponding to the overfitting regime, as confirmed by the eCDFs (Fig.~\ref{fig:membership_attack}). PRIVET was used to assign a score to each sample in the reference data created by merging train and validation sets (Fig. \ref{fig:membership_attack}). The analysis of anomalous NN distances is done, this time, on reference-to-synthetic NN against reference-to-reference NN. We label training samples as positive instances, and validation samples as negative ones. \textcolor{black}{Moreover, a raw nearest-neighbor distance attack defined by thresholding $\log\big( \frac{\delta_S(x)}{\delta_R(x)}\big)$ was added as a baseline, where $x \in R, \quad \delta_S(x) \;=\; d(x, S),$ and $\delta_R(x) \;=\; d\big(x,\; R \setminus \{x\}\big)$.}\\

\cf{In this membership attack experiment we have adapted the PRIVET score~(\ref{eq:new_score}) in two different ways. The first score ($\log_{10} p_r^S$)
consist to replace $p_r^{\rm train}$ by $p_r^{\rm synth}$ where the role of the synthetics is now played by the membership set namely (the members $+$ non-members) and the NN distances are from membership samples to synthetic. The reference metric which was based on synthetic to test NN distance is now from membership to membership. Since the fit is also perform on membership to membership the corresponding probability should 
be $p_{r'}^{\rm memb} = 1/2$ up to small fluctuations so we can discard the denominator and consider simply $\log_{10} p_r^{\rm synth}$. By curiosity we also consider $\Delta p_r(r=1)$ a second derived version of the PRIVET score  which involves $\pi_r$ instead of $p_r$ (order $r=1$)  statistics but keeping the denominator $\pi_r^{\rm memb}$.
What we observe on Figure~\ref{fig:membership_attack} is that the  baseline detect more  members than that  PRIVET score even in the low false positive rate region. The strongest tested statistic is $\log_{10}(\frac{\delta_S}{\delta_R})$. At a false positive rate of $1\%$ it attains a true positive rate of $30.9\%$ against $13.4\%$ for $\Delta p_{r=1}$ and $5.8\%$ for $\log p^S_r$. The high performance of the naive log of distance score is due to a geometric mechanism. By looking at the tail distributions, especially of the members and non-members it is clear that this is made possible by a pattern of overfitting which concentrate first on the isolated training samples having a larger than average NN distance to the other members. A sample lying in a dense region of the reference pool has a small distance to the synthetic set and a small distance to reference pool, whether or not it was trained on, so the raw distance ranks primarily by local configuration and only secondarily by membership.  Stated differently this baseline actually reveals a tendency  that diffusion models tends to overfit first atypical samples and it exploits very well this fact. The local ratio of distances outperforms global $\hat G$ normalization because local density is itself informative.
The second version of the score described above which keeps better track in the denominator 
 $\pi_r^{\rm memb}$ of the distance to the membership set than would do $\Delta\pi_r$ is quite closer and well correlated to the baseline, which confirm the interpretation.
}

\paragraph{\textcolor{black}{Diffusion model}}
\textcolor{black}{The denoiser is a bidirectional Transformer with six pre-normalization Transformer blocks, hidden dimension $256$, eight attention heads, and a feed-forward dimension of $1024$ with GELU activations. Each input is
represented by the sum of a learned token embedding, distinguishing $\{0,1,\mathtt{[MASK]}\}$, and a learned locus-specific embedding. The output consists of one Bernoulli logit per locus, obtained through a final layer normalization and linear projection, with an additional learned locus-specific bias. The model contains approximately $4.95$ million parameters and does not use dropout nor weight decay, the aim being to encourage overfitting. The number of masked loci is sampled uniformly for each training haplotype. Given a corrupted haplotype, the model is optimized using the mean binary cross-entropy over masked loci. Each locus-specific output bias is initialized to the logit of its empirical alternative-allele frequency. The model is trained on all $3{,}500$ training haplotypes using Adam with learning rate $2\times10^{-4}$, batch size $64$, default momentum parameters $(\beta_1,\beta_2)=(0.9,0.999)$, and no learning-rate schedule. Gradients are clipped to a maximum norm of $1$. Synthetic haplotypes are generated using $32$ iterative unmasking steps at temperature $1$. Generation starts
from a fully masked vector.
}
\section{Computational efficiency comparison}
A single A100-48GB was used throughout all this study. \textcolor{black}{We report the runtime of each algorithm for each transform and each fraction of copied samples in the copycat experiment. For PRIVET, the complexity is dominated by computing the 1-NN distances; the EVD fit itself is essentially instantaneous, and computing the $N_{train}$ 1-NN distances within the training set takes less than a minute.}
\label{computational_efficiency}
\begin{table}[h]
\centering
\caption{Computational efficiency comparison (time in seconds, mean ± std) for the copycat experiment on CIFAR10 ($N_{train}=50,000, N_{synth} = 10,000, N_{test} = 10,000$).}
\label{tab:comp-efficiency}
\begin{tabular}{lcc}
\toprule
\textbf{Method} & \textbf{DINOv2 embeddings (768-dim)} & \textbf{Wavelet embeddings (23,232-dim)} \\
\midrule
PRIVET & $9.97 \pm 0.10$ & $10.69 \pm 0.17$ \\
FLD    & $1.01 \pm 0.09$ & $5.62 \pm 0.09$ \\
AUTH   & $0.04 \pm 0.01$ & $1.74 \pm 0.02$ \\
CT     & $8.38 \pm 0.60$ & $105.34 \pm 1.11$ \\
PQM (CPU)    & $333.33 \pm 9.61$ & --- \\
\bottomrule
\end{tabular}
\end{table}

We observed that PQM runs significantly faster on GPU compared to CPU. However, due to memory constraints, it could not be applied to wavelet embeddings, and was not evaluated on DINOv2 embeddings in the current study.

\textcolor{black}{Throughout the experiments, PRIVET's runtime is practical at the scales considered.} Further speedups would be readily obtainable by relying on FAISS library \cite{FAISS_theory, douze2025faisslibrary} which provides exact and approximate nearest-neighbor search. This represents a straightforward avenue for additional scalability improvements, but need to be tested when exact procedures are replaced by fast approximate ones.

\section{\textcolor{black}{Diffusion model trained on a subset of CIFAR10}}
\label{ddpm_cifar10_1k}

\textcolor{black}{For the diffusion model, we train a DDPM on a small subset of CIFAR-10. We first construct a deterministic train/validation split by sampling $1{,}000$ training and $1{,}000$ validation images from the CIFAR-10 training set with a fixed PyTorch generator seed ($\texttt{GLOBAL\_SEED}=43$). The denoising network is a UNet from the \texttt{diffusers} library (\texttt{UNet2DModel}) configured for $32\times 32$ RGB images with channel widths $[64, 128, 256, 512]$, two residual blocks per scale, and self-attention at the deeper layers. The diffusion process uses $T=1000$ timesteps with a cosine noise schedule: we interpolate $\beta_t$ between $10^{-4}$ and $2\cdot 10^{-2}$ on a cosine grid. Training follows the standard noise-prediction objective. We optimize using AdamW with learning rate $2\cdot 10^{-4}$, no weight decay, batch size $32$, and train for $1000$ epochs.}

\section{Impact of embedding on the characterization of underfitting/overfitting regime}
\label{underfitting_dinov2_sec}

\begin{figure*}[!h]
\centering
\includegraphics[width=0.9\textwidth]{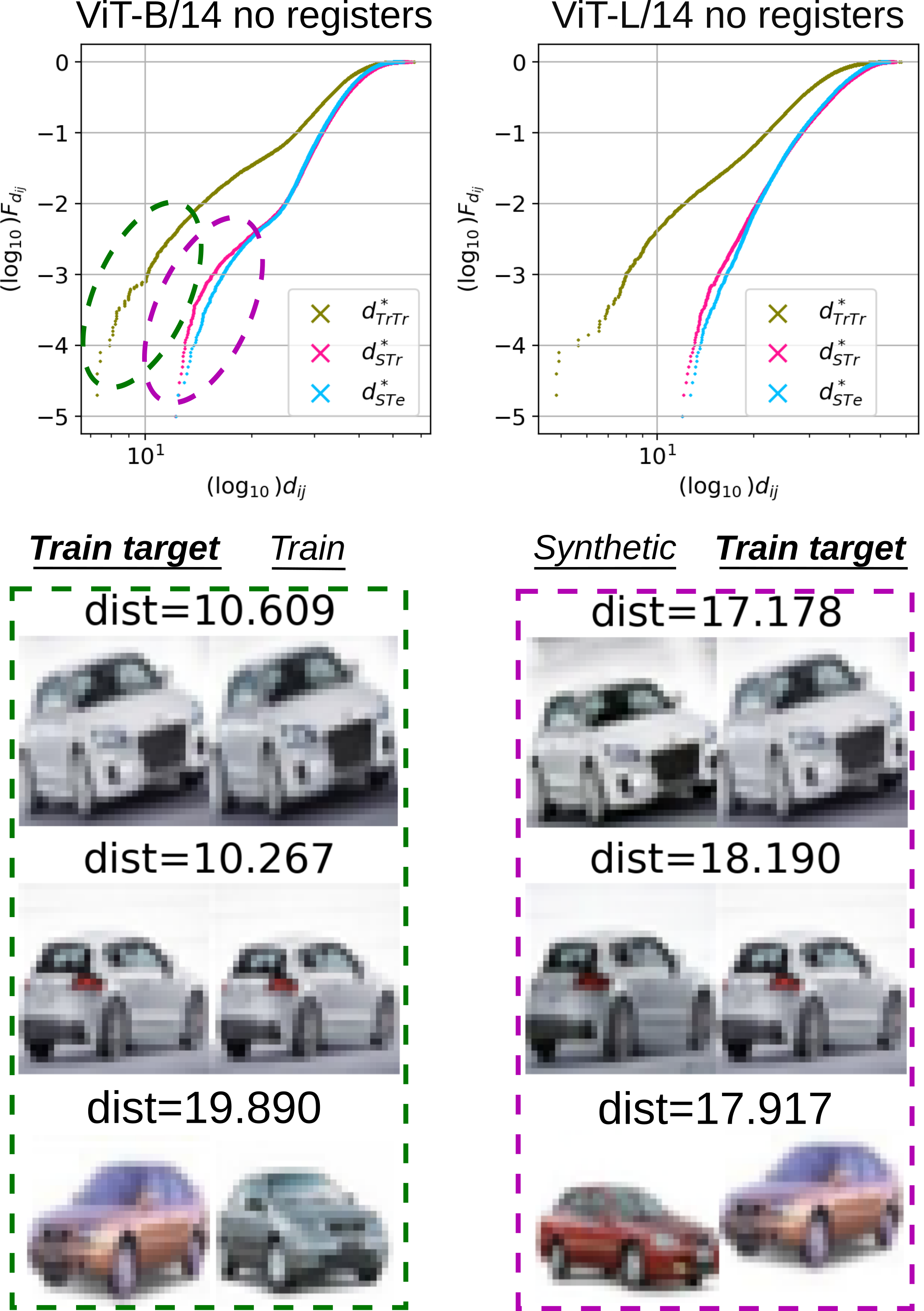}
\caption{\textbf{Synthetic samples generated by PFGMPP appear underfit when using DINOv2 embeddings.} Synthetic samples are generated by PFGMPP diffusion model. \textbf{Top.} Distances are computed on DINOv2 embeddings using ViT-B/14 (86M parameters) without registers (left) and ViT-L/14 (300M parameters) without registers (right). \textcolor{olive}{$d_{TrTr}^*$}, \textcolor{magenta}{$d_{STr}^*$}, and \textcolor{cyan}{$d_{STe}^*$} denote the distributions of nearest-neighbor distances from train to train, synthetic to train, and synthetic to test, respectively. \textbf{Bottom.} We exhibit three pairs of train-train (left column) and synthetic-train (right column), where one training sample is common in both pairs (train target). These pairs appear in the tail of the eCDFs (encircled in color). The distance is the euclidean distance on DINOv2 embeddings.}
\label{app:underfitting_dinov2}
\end{figure*}

Fig.~\ref{app:underfitting_dinov2} highlights synthetic samples generated by the PFGMPP diffusion model that are considered as underfitting when evaluated with DINOv2 embeddings. In particular, examining the tail of the \textcolor{magenta}{$d_{STr}^*$} distribution, we find synthetic samples that are visually nearly identical to their nearest neighbors in the training set, yet their embedding distances are almost twice as large as the distances between that training sample and its 1-NN in the training set (which is a duplicate). A similar case is shown in the second row. In contrast, the third row presents a pair of clearly different images ;  the distance of the (synthetic, real) pair (bottom right) is comparable to that of the nearly identical pairs of the two previous rows, further illustrating the limitations of the embedding when identifying near-duplicate samples. This, together with Main Fig. \ref{fig:copycat_dinov2}, confirm the impact of embedding choice on all distance-based metrics and on the characterization of underfitting and overfitting samples.

\section{Broader impacts}
\label{broader_impact}
The growing use of generative AI for synthetic data generation raises serious privacy concerns, particularly when models are trained on sensitive data such as copyrighted content or human genetic information governed by privacy regulations. Although deep generative modeling has seen rapid progress, the development of rigorous evaluation methods, especially for privacy auditing, has not kept pace. Existing privacy assessments typically rely on global, dataset-level metrics that offer limited interpretability and overlook individual-level risks. In this work, we address this gap by introducing a sample-level privacy metric capable of detecting overfitting and privacy leaks in synthetic datasets. Our method, PRIVET, provides interpretable, fine-grained privacy scores and is applicable across data modalities. We believe this contributes a necessary layer of accountability and transparency to the deployment of generative models, particularly in high-stakes domains like biomedical research and content-sensitive applications.

\clearpage

\end{document}